\documentclass{article}

\usepackage{arxiv}

\usepackage{graphicx}
\usepackage{acronym}
\usepackage{todonotes}
\usepackage{units}
\usepackage{algorithmic}
\usepackage{amsmath}

\usepackage{floatrow}

\usepackage{caption}
\usepackage{subcaption}

\usepackage[hidelinks]{hyperref}

\usepackage{tikz}
\usetikzlibrary{arrows,shapes,calc}
\usepackage{pgfplots}
\pgfplotsset{compat=newest}
\newlength\figH
\newlength\figW
\usepackage{natbib}

\nonstopmode

\title{Estimating Traffic Speeds using Probe Data: A Deep Neural Network Approach}
\author{%
\textbf{Felix Rempe} -  corresponding author\\
	BMW Group\\
	Mobility Technologies \\
	Petuelring 130, 80788 Munich, Germany\\
	felix.rempe@bmw.de\\
	ORCiD: 0000-0002-8007-8152\\
	\And
  \textbf{Philipp Franeck}\\
  BMW Group\\
  Mobility Technologies \\
  Petuelring 130, 80788 Munich, Germany\\
  philipp.franeck@bmw.de \\
  \And
  \textbf{Klaus Bogenberger}\\
  Chair of Traffic Engineering and Control, Technical University of Munich (TUM)\\
  Arcisstrasse 21, 80333 Munich, Germany\\
  klaus.bogenberger@tum.de\\
  ORCiD: 0000-0003-3868-9571
}

\begin{document}
\maketitle

\section{Abstract}
This paper presents a dedicated \ac{DNN} architecture that reconstructs space-time traffic speeds on freeways given sparse data. The \ac{DNN} is constructed in such a way, that it learns heterogeneous congestion patterns using a large dataset of sparse speed data, in particular from probe vehicles. Input to the \ac{DNN} are two equally sized input matrices: one containing raw measurement data, and the other indicates the cells occupied with data. The \ac{DNN}, comprising multiple stacked convolutional layers with an encoding-decoding structure and feed-forward paths, transforms the input into a full matrix of traffic speeds. 

The proposed \ac{DNN} architecture is evaluated with respect to its ability to accurately reconstruct heterogeneous congestion patterns under varying input data sparsity. Therefore, a large set of empirical \ac{FCD} collected on German freeway A9 during two months is utilized. In total, 43 congestion distinct scenarios are observed which comprise moving and stationary congestion patterns. A data augmentation technique is applied to generate input-output samples of the data, which makes the \ac{DNN} shift-invariant as well as capable of managing varying data sparsities. The \ac{DNN} is trained and subsequently applied to sparse data of an unseen congestion scenario. The results show that the \ac{DNN} is able to apply learned patterns, and reconstructs moving as well as stationary congested traffic with high accuracy; even given highly sparse input data. Reconstructed speeds are compared qualitatively and quantitatively with the results of several state-of-the-art methods such as the \ac{ASM}, the \ac{PSM} and a standard \ac{CNN} architecture. As a result, the \ac{DNN} outperforms the other methods significantly.

\hfill\break%
\noindent\textit{Keywords}: Traffic Speed Estimation, Deep learning, Artificial Neural Networks, Floating Car Data, Congestion Patterns

\newpage
\acresetall

\section{Introduction}
Accurate macroscopic traffic speeds on freeways are key inputs for many applications in traffic engineering: travel time prediction, hazard warnings, traffic control, road planning etc. \cite{vanLint.2005,trajectoryRecons.2008,freewaycontrolspeed.2007,HAN2017405,freewayTravelTimeprediction.2001}. Often, traffic speeds are deduced from inductive loop detectors buried in the road surface measuring speeds from passing vehicle at fixed locations \cite{COIFMAN2009349}. The continuous and precise measurements come along with high costs for installation and maintenance of such detectors, which limits the number of units which can be installed. Consequently, with sensor spacings often exceeding several kilometers, measured data is sparse in space. Congestion occurring in between two detector locations may be undetected, or its real impact is estimated imprecisely. 

Thanks to the rapid advancement of mobile technology, other sensor technologies are available, providing valuable information for the accurate estimation of traffic speeds on roads at large scale \cite{Herrera.2010b}. One of them is \ac{GPS} data collected by road users, also called probe data or \ac{FCD}. Given abundant data in a fine spatio-temporal resolution, collected locations and timestamps of vehicles passing a road enable highly precise traffic speed estimations for all times and spaces on a road. Nevertheless, depending on the prevailing penetration rate of reporting vehicles, in practice usually only data from a small subset of vehicles is available. Furthermore, due to technical restrictions of \ac{GPS} and limited sampling frequencies of the \ac{GPS} modules, varying between seconds and minutes, reported speeds are noisy. Additionally, especially in free flow conditions, individual vehicle speeds spread around the macroscopic traffic speed.

An often addressed problem in traffic engineering is the estimation of traffic speeds in space and time given sparse data. The objective is to minimize the error of the estimated traffic conditions compared to the real traffic conditions. Depending on available data and the time interval for which an estimation shall be computed, the problem class is denoted as retrospective traffic speed estimation (also called reconstruction), real-time traffic speed estimation and traffic speed prediction. In a retrospective approach, all measurement data are available and the prevailing traffic state within a time interval in the past is to be determined. In real-time and predictive applications, data up to the current point in time are available and the current or future traffic speed is predicted. The approach presented in this paper, focuses on traffic speed reconstruction. 

In literature, there exist multiple approaches to traffic speed reconstruction, though most of them require that not only speed data but also density or flow measurements are available. Given \ac{GPS} data from only some vehicles driving on the road and, no penetration rate of reporting vehicles, vehicle flow and density can not be inferred. This impedes the application of most common model-based approaches that build upon the \ac{LWR} model \cite{Lighthill.1955,Richards.1956} in order to estimate traffic conditions. Some approaches use an assumed fundamental diagram that relates traffic speed, flow and density in order to estimate missing values \cite{Herrera.2010}. However, the resulting estimated densities scatter significantly around the true density.

Still, there are some approaches that are able to reconstruct traffic speeds using only speed measurements. They can be classified into data-driven approaches based on traffic flow theory, purely data-driven approaches and hybrid approaches. The first category contains models that are motivated by empirical findings of traffic flow dynamics. Similarities and dependencies in collected data samples have been studied and recurrent patterns have been extracted that seek to abstract traffic dynamics. Popular approaches are the \ac{ASM} \cite{Treiber.2003ASM, gasm2010}, the \ac{PSM} \cite{Rempe.2017.PSM}, the ASDA/FOTO model \cite{palmer2011asda}, and others \cite{Work.2010,BekiarisLiberis.2016}. Purely data-driven approaches, comprising the second category, apply generic algorithms to data in order to extract relevant patterns, i.e. to propose a model. Some approaches are based on iterative schemes, that propose, validate and improve the model within multiple computational steps. Hybrid models couple data-driven approaches with information provided by differential equations describing a physical system. Recent approaches include \acp{PINN} \cite{raissi2017physics,YUAN202188}. 

Recently, deep learning has gained much attention as one data-driven machine learning technique. Corresponding approaches have been successfully applied to problems in image classification, speech analysis, robotics, etc., outperforming classical approaches significantly \cite{resnet, LeCun2015}. Also in traffic engineering, (deep) learning techniques have been applied successfully to some problems. Especially for predictions, there are many approaches, e.g. \cite{cnnprediction,vanLint.2005,lana.prediction, Vlahogianni.2013,deepprediction}. Though, most of these approaches focus on the prediction of speeds, travel times or flows based on complete current and historical data.

Due to the success stories of \acp{DNN} applied to various other data-centric problems \cite{Dargan2019ASO}, it is a promising technique for traffic speed estimation as well. It rises the expectation that the automated learning process is able to extract the dynamics of traffic flow from data solely. Subsequently, a fully trained \ac{DNN} may be able to infer complete traffic speeds based on learned traffic patterns, given only a few measurements.

Literature on neural networks applied to sparse speed data are mostly trained and studied with data observed on a short road stretch and short time interval only. Data used for training often contains few distinct congestion patterns, such as minor shockwaves. It is an open question, whether these approaches are able to distinguish between varying congestion patterns and reconstruct those. Also, many published network architecture are designed to solve the traffic speed estimation problem for a fixed space-time domain. Since the affected road length as well as the duration of empirical congested traffic varies significantly, these methods lack flexibility to be applied at large scale. Another requirement for large scale application is that data of a single sparse data source is sufficient for model training and inference. Additionally, The estimation method shall be robust against a varying input data density, which is the case especially given probe data.

This paper presents a dedicated \ac{DNN} architecture that adapts to the needs and potentials of processing real probe data. The learning capabilities of the \ac{DNN} are studied using a large amount of probe data collected during 43 congestion scenarios on a motorway. A dedicated \ac{DNN} architecture is presented, that can be trained given sparse and noisy speed data of a single source. The presented \ac{DNN} is derived from the U-net \cite{ronneberger2015unet} and comprises multiple units of convolutional layers, an encoding-decoding structure and feed forward paths. Instead of learning congestion patterns on large space-time domains, which requires vast amounts of data and processing time, the idea is to break town the problem of reconstructing a complex congestion pattern into several small tasks. Therefore, space-time domains are decomposed into smaller matrices, and training and inference is done on these small matrices. This allows the network to generalize the reconstruction problem and learn patterns easier. In addition to using only a speed matrix as input, an occupancy matrix indicating the existence of measured data in input cells is fed to the network. This trains the network in handling varying data densities. 

For evaluation, the developed network is trained using part of the large dataset and subsequently applied to data of an \textit{unseen} complex congestion scenario comprising moving and stationary congestion. The results show that, despite sparsity of the data, the \ac{DNN} is able to reconstruct these patterns accurately. In a comparison with state-of-the-art approaches such as the \ac{ASM}, \ac{PSM} and a standard \ac{CNN}, the estimation accuracy under varying data density is studied. The results show, that the \ac{DNN} outperforms the other approaches significantly.

The paper is structured as following: Section \ref{sec:sota} summarizes related work and emphasizes the contribution of this paper. Section \ref{sec:notation} introduces available data and notation. Section \ref{sec:trafnet} motivates the applied deep convolutional neural network architecture that turns sparse \ac{FCD} into a continuous speed estimate. Section \ref{sec:eval} provides an analysis of the trained network and gives insights into the learned state. A quantitative comparison with state-of-the-art methods, such as a simple smoothing approach, the \ac{ASM}, the \ac{PSM} and a standard \ac{CNN} is given in section \ref{sec:evaluation}. Sections \ref{sec:discussion} and \ref{sec:conclusion} discuss the findings and give an outlook.

\section{Related work} \label{sec:sota}
The main challenge of reconstructing traffic speeds from \ac{FCD} is that this data is usually sparse in space and time. If one is interested in the average speed of traffic prevailing at position $x$ along a road at time $t$. In some cases, there might be data of multiple vehicle trajectories in close temporal and spatial proximity available, such that the average traffic speed can be inferred with high certainty using space-time averages. In other cases, collected data may be minutes or kilometers apart from the relevant time-space location $(t,x)$. Since traffic conditions change dynamically over time and space, it is uncertain how collected data can be utilized to estimate the speed at $(t,x)$. In this case, a traffic model may provide valuable information since it encodes common patterns of traffic dynamics. Thus, a model calibrated with measured data is potentially able to estimate conditions more accurately.

Approaches that deal with traffic speed reconstruction on freeways can be divided into four categories: (i) analytical macroscopic traffic flow model relating density, flow, and speed; (ii) data-driven models based on empirical traffic flow theory; (iii) purely data-driven methods and (iv) hybrid models that combine two or more approaches. To narrow down related work, the focus of the review is on approaches relying primarily on probe data.

Approaches that belong to the first category are usually based on the \ac{LWR} model \cite{Lighthill.1955,Richards.1956}. They combine model predictions with measurement data using e.g. a Kalman filter \cite{vanLint.2014,Suzuki.2003}. Also, higher order models that are supposed to model traffic dynamics more realistically are chosen as base model \cite{Aw.2000,Wang.2005}. While the mentioned approaches rely on flow data collected by loops, recent approaches strive towards \ac{GPS} data. \cite{Herrera.2010,Herrera.2010b,Work.,Work.2010,Work.b} derive models based on the \ac{LWR} which is updated using speed data only. Nevertheless, complete boundary conditions and a \ac{FD} are required. \cite{BekiarisLiberis.2016} describes an approach that requires the density of connected vehicles as well as their speeds in order to estimate the traffic state using a second-order \ac{LWR} model. However, also in this approach the boundary conditions need to be known in advance.

The second category comprises methods that are based on empirical traffic flow theory without using a \ac{PDE} of traffic dynamics. The well-known \ac{ASM} makes use of the characteristic speeds of shock waves in free and congested traffic: in congested traffic shockwaves reportedly propagate upstream with $v_{cong} \approx \unit[15]{km/h}$ \cite{Treiber.2013,Kerner.2004}. In free flow shockwaves typically propagate downstream with $v_{free} \approx \unit[80]{km/h}$ \cite{Treiber.2013}. The \ac{ASM} smooths traffic speed data using convolutional filters in space and time that consider the propagation speeds \cite{Treiber.2003ASM}. Originally invented to reconstruct traffic conditions from stationary loop detectors, the \ac{ASM} has been successfully applied to \ac{FCD} and other data sources \cite{gasm2010,vanLint.2009,Rempe.MITITS,Rempe.2016GASM}. Inspired by the three-phase traffic theory, which distinguishes between two congested phases --- the synchronized traffic flow and \acp{WMJ} \cite{Kerner.1999,Kerner.2004} --- two further models are the ASDA/FOTO and \ac{PSM}. The ASDA/FOTO model utilizes sensor data to track the fronts of traffic phases \cite{Kerner.2004b}. The original model was developed for loop data and further adapted in \cite{palmer2011asda} to be applied to probe data solely. The \ac{PSM} expands the \ac{ASM} considering that synchronized congestion patterns are often localized \cite{rempe2017phase}. Applied to \ac{FCD} only, this results in higher reconstruction accuracies.

The third category comprises data-driven approaches that are agnostic to traffic flow dynamics. Characteristic of these is that they are in principle applicable to a variety of problems and that they are calibrated manually or automatically to estimate traffic speeds with sparse data. Please note that the focus is the retrospective estimation of traffic speeds. For a more comprehensive overview of traffic speed prediction approaches, please refer to \cite{Vlahogianni.2014,lana.prediction}. In literature, several approaches exist that intend to estimate traffic conditions retrospectively based on sparse data. \cite{CokrigingImputation} apply a Kriging based model that imputes missing data using spatio-temporal dependencies of nearby sensor data. \cite{TensorBasedCompletion} propose a tensor-based method that exploits correlations between data and allows reconstructing missing historical data. \cite{TensorBasedCompletion2} also rely on tensors, estimating missing volume data using spatial correlations. \cite{DataImputationTensor} apply a tensor-based approach targeting the imputation of traffic speeds in time-series. In recent years, also neural networks have been applied to reconstruct missing traffic data. \cite{ImputationMultiViewLearning} utilize a \ac{LSTM} combined with \ac{SVR} to impute missing loop data. Time series of traffic flow are imputed and predicted with a \ac{DNN} in \cite{DeepLearningImputation}. \cite{DeepLImputationAppl} use an encoder-decoder structured neural network to learn latent features for combining noisy and erroneous traffic data. \cite{DataImputationCNN} reconstruct traffic volume data using a \ac{CNN}.

The later mentioned approaches \cite{CokrigingImputation,TensorBasedCompletion,TensorBasedCompletion2,DataImputationTensor,ImputationMultiViewLearning,DeepLearningImputation,DeepLImputationAppl,DataImputationCNN} propose purely data-driven methods, and some of them are deep learning methods. However, they mostly rely on loop data and are designed to impute, for example, traffic volumes due to sensor errors. The proposed methods are not easily applicable to speed data such as \ac{FCD} that is collected at varying times and places. A few recent approaches study the application of neural networks using probe data: \cite{deepCNNtrafficdata} propose a \ac{CNN} model with an encoding-decoding structure and train it on simulated data and NGSIM data. The results are promising as the network outperforms state-of-the-art methods such as the \ac{ASM}, nevertheless, the space and time interval of NGSIM data is quite limited, and among all empirically observed congestion patterns \cite{Helbing.2009,Kerner.1999} in NGSIM dataset mostly shockwaves can be observed. For instance, stationary congestion patterns are missing. Another approach is described in \cite{thodi2021incorporating}. Compared to conventional \acp{CNN}, the authors formulate constraints that force the parameters of the \ac{CNN} layers to convolve data with shockwave-characteristic speeds. Also this approach is trained and evaluated with NGSIM data.  

A recent class of algorithms that belong to the fourth category, i.e. hybrid models, are \acp{PINN}. They extend the purely data-driven methods with an analytical traffic flow model, e.g. the \ac{LWR} model. The idea is to assess the computed traffic state of a neural network not only based on measurement data, but also on its compliance with the given physical model. Recent approaches include \cite{pinn_agarwal}, who publish an approach giving promising insights into the ability to reconstruct traffic densities based on sparse measurements; \cite{liu2020learningbased}, who use simulated trajectory data providing local densities, study an approach to reconstruct the traffic density in space and time and \cite{shi2021physicsinformed}, who design a \ac{PINN} with additional error term based on the \ac{LWR} and a Greenshields \ac{FD} in order to estimate traffic density on a road. These approaches show promising results, though, published network architectures are not applicable to the problem of traffic speed estimation with probe data: Some approach require that probe vehicles provide density information \cite{liu2020learningbased}, which is rarely the case. Additionally, the network needs to be re-trained to fit the data in each situation \cite{shi2021physicsinformed}. This is computationally expensive, and is prone to overfitting since the network's output is optimized to match observed data.

This paper presents an adapted \ac{DNN} architecture that builds upon the well-known U-net \cite{ronneberger2015unet} to meet the special requirements of a traffic speed estimator using sparse probe data. In comparison to existing approaches, the \ac{DNN} presented in this paper applies a concept to decompose the space-time domain into regular smaller matrices in order to reconstruct congestion scenarios of variable temporal and spatial extent. Furthermore, in this approach, training and evaluation is based on a large data set comprising empirical probe data collected during 43 congestion scenarios. Due to the amount of data, the neural network architecture is able to learn various, heterogeneous congestion patterns that occur in real traffic. Due to the variety of congestion scenarios and the naturally varying data density, the approach is less prone to overfitting compared to networks trained on data collected on a short road stretch only (NGSIM) or using simulated data. Additionally, the temporal and spatial extent of studied congestion situations exceed the data given in NGSIM, which is the basis for recent approaches, significantly, which allows to study a \ac{DNN} on a larger time and space domain. Finally, the presented approach is robust to noisy and incomplete data under varying data density in time and space. This promises a quick application to related problems.

\section{Notation \& Data} \label{sec:notation}
On a road stretch with length $X$, all raw measurements within a time interval $[T_0, T_0+T]$ are considered. For further processing, time and space are discretized into a uniform grid with cell sizes $\Delta T \times \Delta X$ and a total of $n_T$ cells in the time dimension, and $n_X$ cells in the space dimension. The speed in each cell is considered constant and denoted as $v_{i,j}$ where $i = 1, \dots, n_T$ and $j=1,\dots, n_X$. 

The sensors of an equipped vehicle sample \ac{GPS} with a certain frequency. Timestamps and positions can be mapped onto the road stretch and are available as tuples $(t,x)$. In-between two sampled locations, the vehicle's speed is assumed to be constant as there is no further information. For each grid cell which is passed by the vehicle, i.e. the vehicle traveled $\Delta x_{i,j} \ge \unit[0]{m}$ and $\Delta t_{i,j} > \unit[0]{s}$ in that cell, the cell-wise speed is determined as $v_{i,j} =  \Delta x_{i,j}/ \Delta t_{i,j}$. All cell-wise speeds of a trace are computed, and subsequently, the speeds of all traces are written into a sparse matrix $V_{FCD} \in \mathbf{R}_+^{n_T \times n_X}$ comprised of tuples $(i,j,v)$. If there are multiple speed entries from different trajectories for the same cell, the harmonic mean of all speed values assigned to that cell is considered. Figure \ref{fig:patch} a) depicts schematically two traces and the resulting assigned speed cells.

\section{Feature Setup \& Network Architecture} \label{sec:trafnet}
In order to process sparse speed data into a complete and accurate estimate of traffic speeds using a neural network approach, some preliminary considerations are made. The challenge is to design the input and expected output in such a way that the neural network is able to learn the input-output relation and be able to apply the mapping to unseen data in an optimal way. Given a well-fitting setup and neural network architecture, a neural network is able to generalize the problem without overfitting \cite{goodfellow2016deep}. The input to the neural network needs to contain all relevant information for solving the problem. At the same time, too much information makes learning more difficult.

The chosen approaches presented in this paper make use of recent advances in image processing using \acp{DNN}. State-of-the-art methods apply multiple layers of convolutional neural networks to pixels of an image \cite{goodfellow2016deep,long2015fully,resnet}. In that way, the relation between the intensities of nearby pixels is evaluated within the first convolutional layers of the neural network, and dependencies on a larger scale are considered in later neural network layers.

The challenge of reconstructing traffic conditions from sparse data has many similarities to image processing, but also some differences. The speeds represented in a uniform two-dimensional grid are reminiscent of the pixels of an image, and it is clear that traffic conditions at space-times $(t,x)$ are related to the traffic conditions in close space-time proximity, since traffic dynamics are non-instantaneous, i.e. it takes time until congestion on a road forms and dissolves. Therefore, as in image processing, convolutional layers appear to be a good choice to capture the interdependency of traffic conditions in proximate grid cells. One difference, though, is that, while image processing usually aims at the classification of images (e.g. whether there is a cat in a picture), in this case, a continuous output is required, which is called a regression network \cite{goodfellow2016deep}. Another difference is that the given input data is sparse in time and space. This is a problem rarely studied with neural networks.

\subsection{Input and output features} \label{sec:inout}
\begin{figure}[tbh]
	\centering
	\includegraphics[width=.98\textwidth]{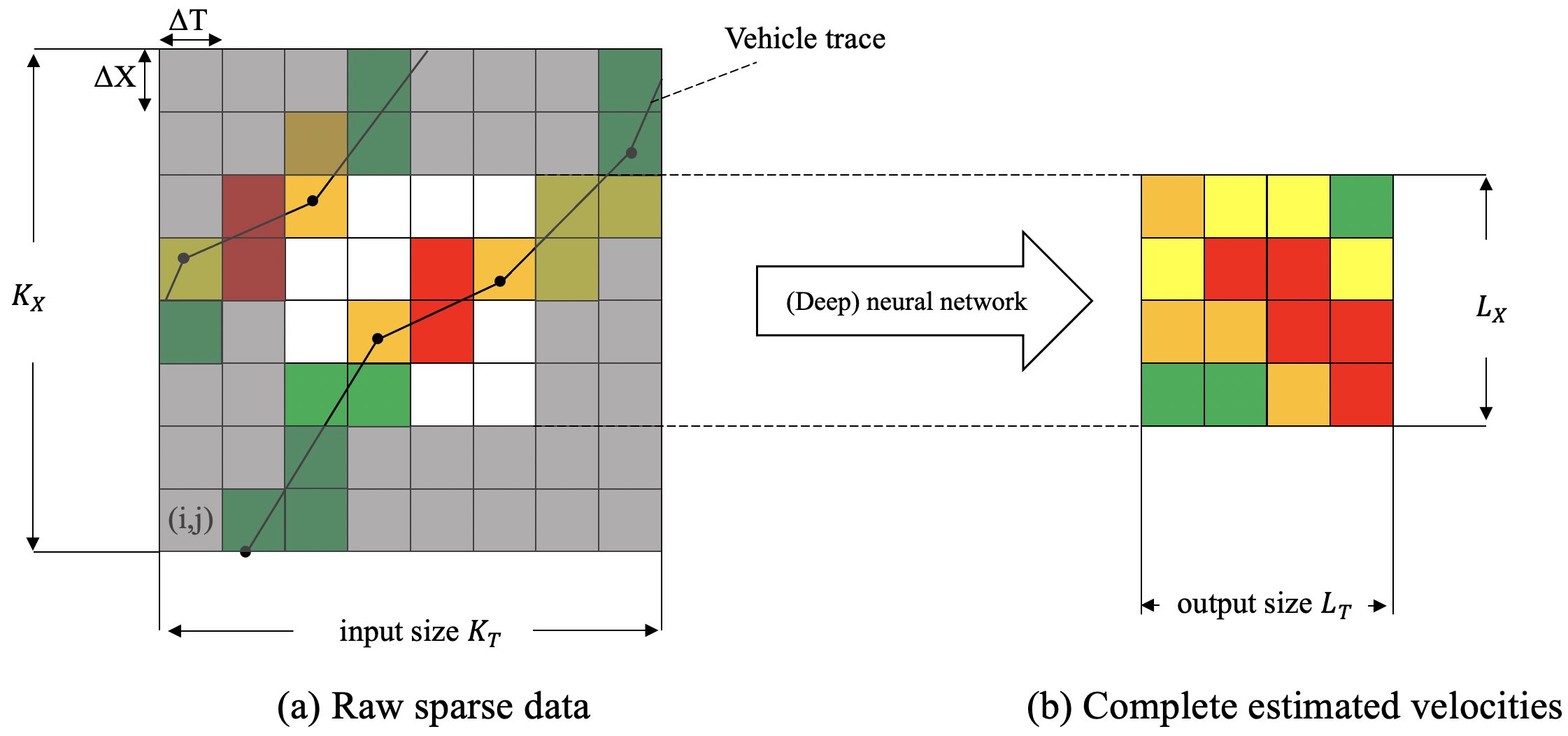}
	\caption{Schematic representation of trajectory data that is converted into cell speeds of a patch. The colors depict the vehicles' speeds passing a grid cell. The black dots in a) represent the locations and times at which GPS data is sampled. The sparse data of a patch is processed using a (deep) neural network approach. The inner cells (on the right) are considered as output speeds.}
	\label{fig:patch}
\end{figure}

Since a congestion scenario may vary with respect to the affected road stretch and the time interval, the reconstruction task needs to be generalized such that any situation can be reconstructed. Therefore, the complete domain is subdivided into smaller matrices, called patches in this context (see Figure \ref{fig:patch}). A patch $V_{in}$ is a $K_T \times K_X$ matrix containing the sparse data of $V_{FCD}$ within the respective bounds. The output of the method is another, but smaller matrix (or patch) with $L_T \times L_X$ cells. The difference $K_T - L_T$ and $K_X - L_X$ are the patch boundary sizes. They allow the configuration of the minimum temporal and spatial proximity that is considered in order to estimate traffic conditions in cell $i,j$. 

Congested traffic that occurs within a space-time domain usually exceeds the size of an input patch. For the reconstruction of the entire event, the domain is subdivided into multiple input patches. Two neighboring patches have an overlap of $K_T - L_T$ cells in time dimension, and $K_X - L_X$ cells in space dimension, such that the resulting output patches can be stitched together without any gaps. In order to ensure that the total output domain size equals the input size, the input domain is padded with empty grid cells on all boundaries with $K_T-L_T$ and $_X -L_X$, respectively.

\subsection{Neural network architecture} 
\begin{figure}[tbh]
	\centering
	\includegraphics[width=.98\textwidth]{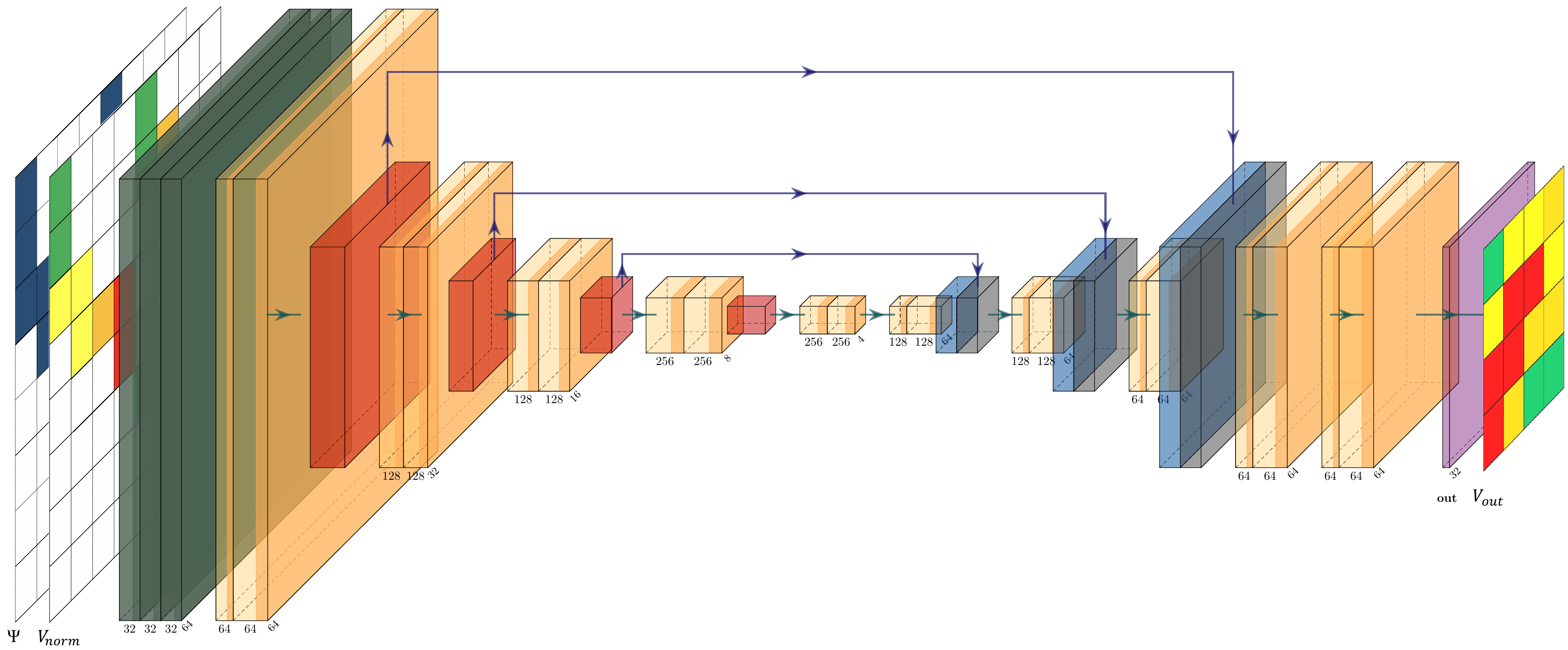}
	\caption{The applied \ac{DNN}. Colors depict different types of layers: (1) dark green: 3D convolution layer with \ac{ReLU} activation and batch normalization; (2) light orange: 2D-convolutional layer with ReLU activation and batch normalization; (3) dark orange: 2D max-pooling layer with stride (2, 2); (4) blue: 2D upsampling layer; (5) gray: concatenation layer; (6) purple: 2D convolution layer without activation. Directly below each layer, the number of filters of that layer is given. The layer dimension is noted in diagonal letters. Note: Although input matrices $\Psi$ and $V_{norm}$ are full matrices, for reasons of clarity, they are visualized as sparse matrices only showing assigned cell data and with reduced size, in this schematic representation.}
	\label{fig:unet}
\end{figure}

Finding the hyper parameters of a neural network that solves the reconstruction problem most accurately is the second important step \cite{goodfellow2016deep}. After many experiments, we present the final neural network architecture that resulted in the best results (see section \ref{sec:eval}).

The \ac{DNN}'s architecture is depicted in Figure \ref{fig:unet}. It is based on the U-net, which was originally invented for image segmentation in medical imaging \cite{ronneberger2015unet}. It is a fully convolutional neural network with an encoding-decoding structure \cite{long2015fully}. Based on the good experiences in an application to the reconstruction of missing data in seismic and medical applications \cite{Chai2020,QIMS29735}, the basic idea is adopted here, and the network architecture is fit to the traffic reconstruction problem. 

Input to the neural network are two matrices. The first comprises normalized speed data, containing normalized speeds $v_{norm}$ that are deduced from sparse input $V_{in}$. A normalization step usually simplifies the regression task for a neural network. The expected maximum output speed of the neural network is $v \le \unit[130]{km/h}$. For this reason, input values are shifted by $v_{shift} = \unit[65]{km/h}$ and divided by a factor of $v_{var} = \unit[100]{km/h}$:
\begin{equation}
  v_{norm} = \frac{(v_{raw} -v_{shift})}{v_{var}}  
\end{equation}
\noindent Matrix entries without data are set to a zero value in order to fulfill the requirement of a layer to only contain numerical values. 

With the required imputation of values at prior undefined cells, the sparse matrix turns into a full matrix. Given the full matrix, it is not possible to differentiate whether a cell contains a measured speed or a replacement value. In order to provide the neural network this information, a second input matrix is defined. We denote it as the \textit{occupancy} matrix $\Psi \in[0,1]^{K_T \times K_X}$. It contains a value of one at $(i,j)$ if the input patch matrix $V_{in}$ contains a valid measurement in that cell, and zero otherwise:

\begin{equation}
  \Psi^{i,j} = \begin{cases}
    1 \text{ if } (i,j) \in V_{in} \\
    0 \text{ else .}
  \end{cases}
 \end{equation}

\noindent 
The normalized speed matrix, and the occupancy matrix are concatenated along a third dimension, such that the input space becomes three-dimensional. The first layers of the network comprise several 3D convolutional layers with a filter size of $5\times5\times2$ and a \ac{ReLU}. These 3D convolutional filters work on raw data and occupancy data simultaneously. A subsequent batch normalization layer reduces the covariance shift of the output applying a normalization step \cite{batchnorm}. This layer is applied based on the good experiences of other DNN approaches which report enhanced robustness and accuracy of the trained network. Layer 10, a 3D MaxPooling step, finally aggregates data along the third dimension, effectively reducing the intermediary data dimension to two.

Several blocks of 2D convolutional layers with \ac{ReLU} activation and 2D MaxPooling are applied (comparable to the original U-net paper \cite{ronneberger2015unet}). With decreasing output size, the number of filters increases in order to store the latent information of this encoding scheme. The subsequent decoding part comprises stacked units of 2D Upsampling layers with bilinear upsampling technique, a concatenation layer which is an \textit{identity shortcut} of same-sized previous outputs, 2D convolution layers and a batch normalization. This shortcut turned out to enable faster convergence and more accurate results in the U-net \cite{ronneberger2015unet} and other state-of-the-art deep networks \cite{resnet}.

When the final output dimension of $L_T \times L_X$ is reached in the decoding path, four additional convolutional layers are added that give the neural network the power to further transform data from the decoding sequence. A final 2D convolutional layer reduces the channels to one output value per cell. No activation function is needed since a continuous (normalized) output is expected.

\begin{figure}[t]
	\centering
	\includegraphics[height=.98\textwidth]{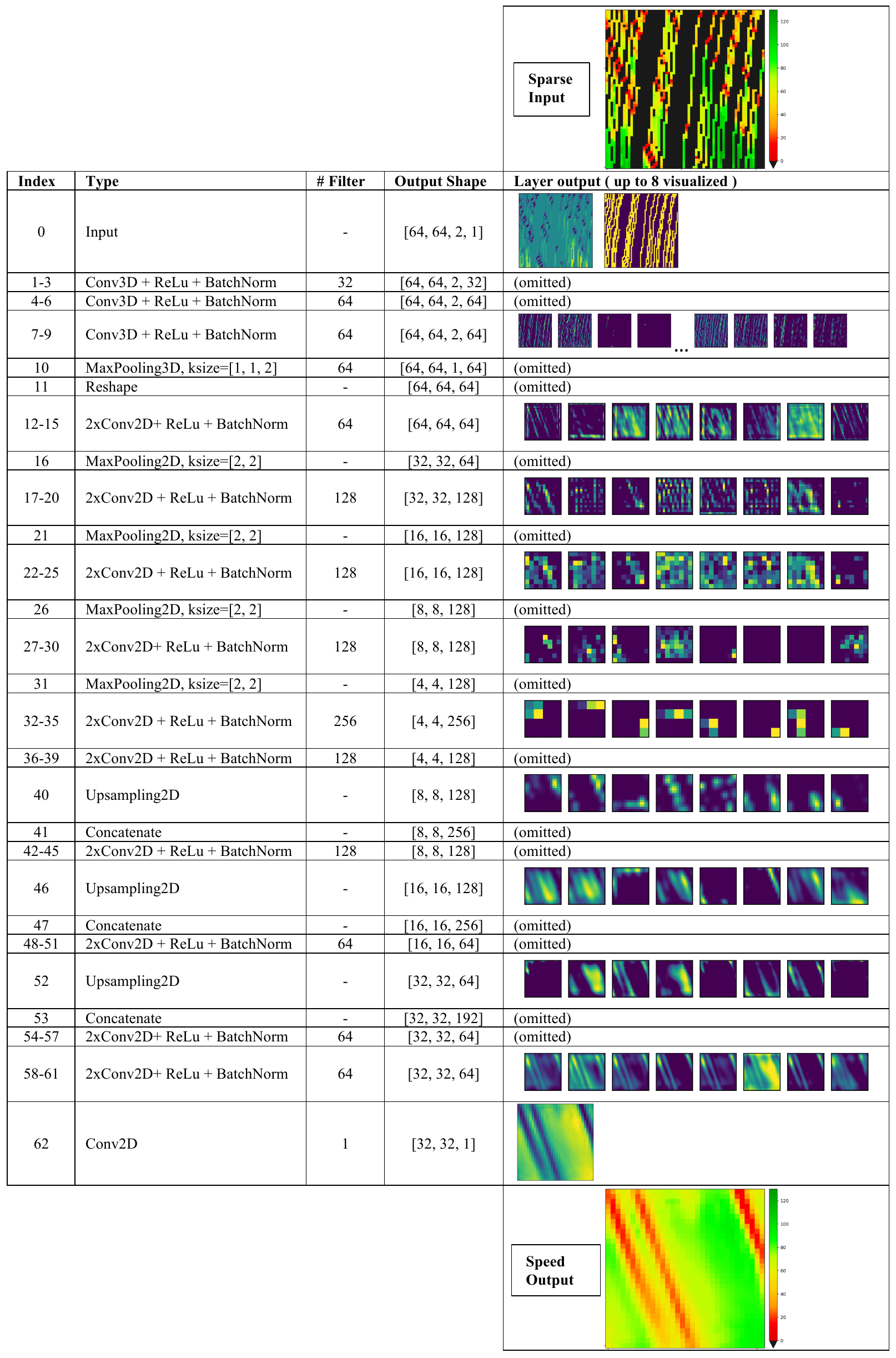}
	\caption{The 62 layers of the \ac{DNN} in detail, along with up to 8 output matrices produced by various layers given one example input patch. Note that the input patch size is larger than the output patch in order to avoid reconstruction artifacts at patch boundaries.}
	\label{fig:layer_details}
\end{figure}

The exact layer definition is given in Figure \ref{fig:layer_details}. All in all, the network has 8,506,017 trainable parameters. In addition to the layer details, the layer outputs of fully trained network (see section \ref{sec:training}) given an example input patch are visualized. In the following, the neural network is called \textit{TraNet} for easier reference. 

One may wonder whether a network with fewer layers might not solve the problem as well. In fact, we started experimenting with multiple standard \ac{CNN} networks in order to solve the problem of traffic estimation using sparse data. However, none of them turned out to come close to the performance of the reference methods \ac{ASM} and \ac{PSM}. It appeared, that, in order to reconstruct spatio-temporal congestion patterns, data in grid cells, that are far away in space and time, need to be related. To capture the distant spatio-temporal dependencies, one can either apply large convolution filters with many parameters, and stack a few of them, or, apply small filters and stack many of them. At the same time, all empty grid cells need to be ignored, since they do not contain information. An simple approach would be to downsample the input space in order to reduce the number of cells and, likewise, the required filter size. However, this reduces the resolution of the data, and valuable information is lost. As it turned out during the experiments, the training with a large number of small filters was more successful than with a few large filters, finally resulting in the presented deep neural network. 

\clearpage

\section{Evaluation} \label{sec:eval}
The evaluation of the approach is structured in the following way: First, study site and available data are summarized. Second, the error metric and training methodology of the neural network are described. Third, the neural network is applied to one congestion scenario with varying input data and insights of the learned layers are given. Finally, a quantitative comparison with state-of-the-art methods is performed that contrasts the strengths and weaknesses of the neural network approach.

\subsection{Data}
\begin{figure}[t]
	\includegraphics[width=.9\textwidth]{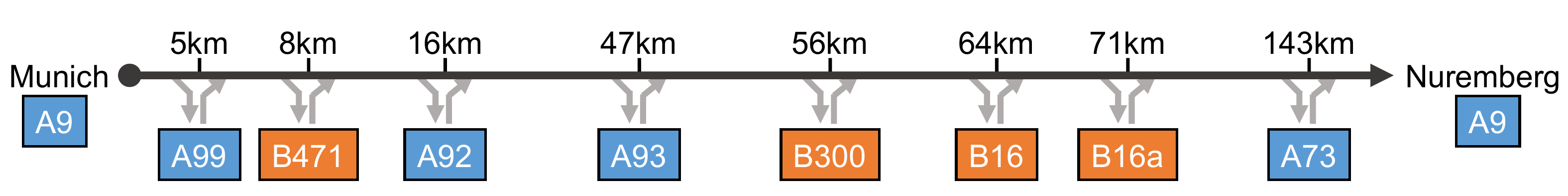}
	\caption{Intersections of Autobahn A9 with other freeways and major roads.}
	\label{fig:mapOfA9}
\end{figure}

As the study site, the German Autobahn A9 is selected (see Figure \ref{fig:mapOfA9}). \ac{GPS} data sampled by a fleet of vehicles is collected and map-matched onto the road geometry of the digital map. The vehicles sample their position in time intervals between \unit[5]{s} and \unit[20]{s}, depending on their software version. Collected data is transmitted to a central server, using an anonymous identifier for reasons of data privacy.

\begin{figure}[t]
    
  \includegraphics[width=.99\textwidth]{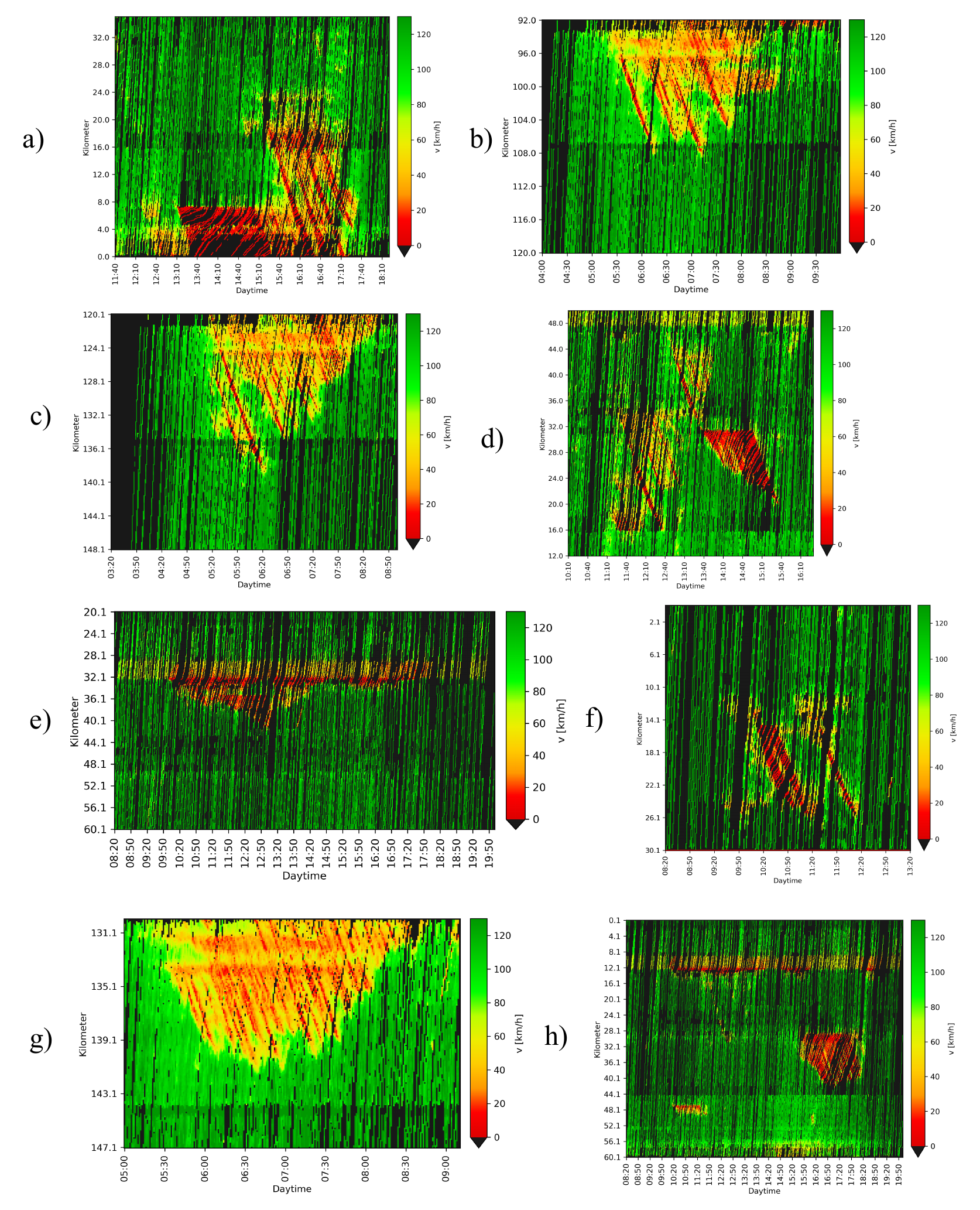}
    {\caption{\ac{FCD} collected during 8 of the overall 43 congestion scenarios in both driving directions}\label{fig:vall_examples} }
\end{figure}

Given the size of the collecting fleet and the high spatio-temporal resolution of the data, it is possible to study the spatio-temporal dynamics of traffic speeds. Figure \ref{fig:vall_examples} visualizes example data collected during congestions on the A9 in the southbound as well as the northbound direction. As described in section \ref{sec:notation}, positions and timestamps of the data are used to compute speeds and assign them to grid cells. Here, a grid size of $\Delta T = \unit[60]{s}$ and $\Delta X = \unit[100]{m}$ is used. All in all, 43 such scenarios collected in November and December 2019 are used for subsequent training and evaluation of the approach.

\subsection{Metric} \label{sec:metric}

In order to assess the accuracy of an estimate, all data is divided into a training (i.e. the input data) and a test set. The training set is used to reconstruct the speeds and the test data is used to assess the reconstructed speeds. The ratio of number of training traces with respect to the total number of traces per scenario is denoted as $p \in [0,1]$. As with the training traces, the test traces are converted into cell-wise speed tuples $(t,x,v)$, aggregated cell-wise and collected in the set $V_{GT}$. Please note that $V_{GT}$ do not necessarily represent macroscopic traffic speeds since a cell speed is deduced from only one or a few of all vehicles that passed the cell. Thus, the values might deviate slightly from the macroscopic speed. However, in congested traffic conditions, where vehicle speeds are synchronized \cite{Kerner.1999}, this deviation is expected to be small. 

In order to assess $V_E$ with $V_{GT}$ a metric needs to be chosen. The metric is a fundamental design parameter of the setup since it constitutes the optimization goal of the network during training. Often, the $MAE$ or $RMSE$ are used. In free flow conditions vehicle speeds are high, and, due to varying travel speeds of individual vehicles in unbounded traffic flow, absolute speeds of nearby vehicles often differ significantly. In congested traffic speed differences are much smaller. Thus, assuming an estimated average traffic speed of any method, there is a naturally large absolute error in free flow traffic compared to the error in congested traffic. Using a metric such as the $MAE$ for training purpose of a neural network causes the optimization strategy to minimize significant errors first, which are by definition wrongly estimated speeds in free flow conditions. 

The anticipated application of this network, however, is the estimation of traffic speeds in potentially congested traffic. For a metric to be deemed appropriate it should be more sensitive in the lower speed ranges, since low speeds impact travel times more severely. Additionally, it should be symmetric, i.e. wrongly over- and underestimated speeds should be penalized equally. A simple metric that fulfills these requirements is the \ac{IMAE}:

  \begin{equation} \label{eq:ASM-v}
  IMAE = \frac{1}{ | V_{GT}  | }  \sum_{(t,x,v)}^{ V_{GT} }   \big|  \frac{1}{V_E(t,x)} -  \frac{1}{ v} \big|   
  \end{equation}
where $|.|$ denominates the cardinality of a set.

\subsection{Training} \label{sec:training}

\begin{figure}[tbh]
	\includegraphics[width=1.0\textwidth]{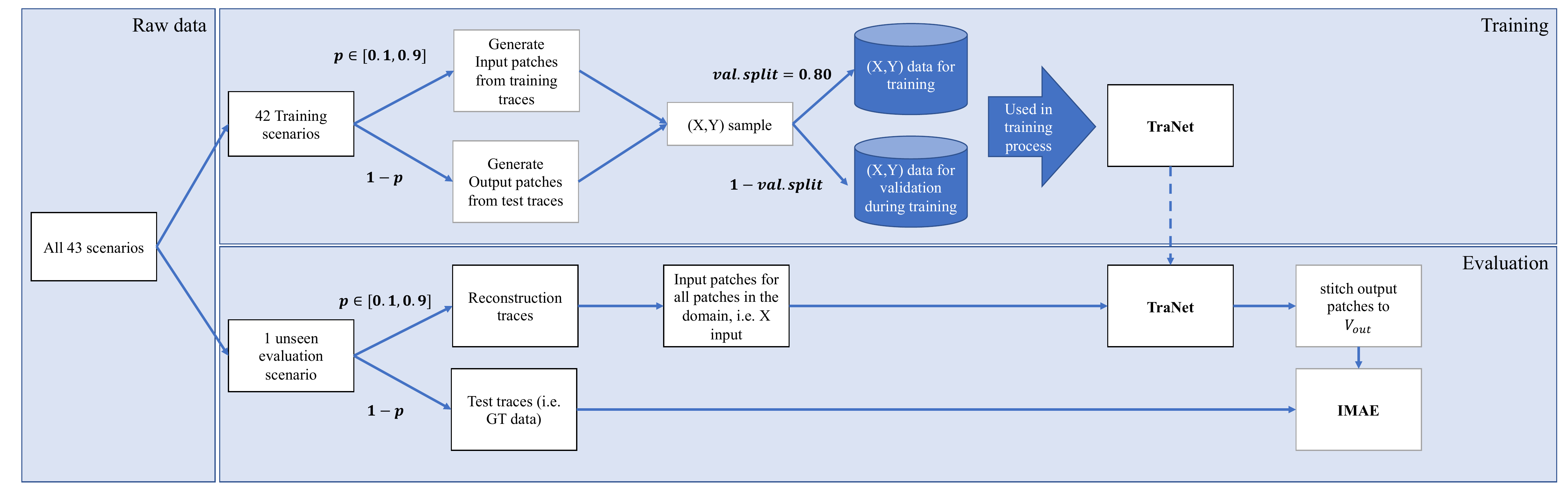}
	\caption{Information flow during training and inference stage using \ac{TraNet}.}
	\label{fig:traintestflow}
\end{figure}

Training the neural network requires a large number of input-output samples which should cover as many different situations as possible. For this study, 43 congestion scenarios are available in total. A straightforward approach would be to uniformly divide all scenarios into patches and use these patches as training data. However, the number of resulting patches would be too low to allow the neural network to generalize the reconstruction problem, since the distribution of assigned speed cells of an unseen patch may be disparate to the patches used for training. Furthermore, the reconstruction accuracy is sensitive to the prevailing data density \cite{Rempe.2017.PSM,Rempe.2016GASM,gasm2010}. Therefore, in this case a data augmentation approach for the generation of patches is chosen (see Figure \ref{fig:traintestflow}): Randomly, one of the 42 congestion scenarios used for training is picked and from all available trajectories a set of training traces is drawn from all available trajectories with a ratio of $p\in [0.1,0.9]$. These traces are used to generate input patches at random locations of the scenario's space-time domain. Accordingly, the remaining test traces are converted into matching output patches for the same space and time as the input patches. Note that the output patches are sparse as well. The set of speeds contained in the output matrix are used as $V_{GT}$ for computing the \ac{IMAE} during training. Further note, that the speeds used as \ac{GT} data do not necessarily resemble the macroscopic traffic speeds. Rather, they represent samples that spread around the true macroscopic speed. Nonetheless, this approach has the significant advantage, that no complete \ac{GT} data is required to train the network, instead, sparse data of a single source is sufficient. 

All generated input-output patches are collected, and the procedure is repeated until a total number of patches $N_{in}=N_{out}=100000$ is available. In this way, a huge set of samples that cover all possible congestion patterns with varying input data density are created. Although the number of original congestion scenarios appears to be small at first glance, each scenario itself contains various congestion patterns. With the random deduction of patches from the original data, it is accomplished that the neural network learns to be \textit{shift invariant} \cite{zhang_shiftinvariance} and is able to handle variable data density. 

As depicted in Figure \ref{fig:traintestflow}, the congestion scenario that is supposed to be reconstructed, is completely unseen to the network. For later evaluation, available data of this scenario is split into a reconstruction and an evaluation set, using the reconstruction and \ac{TraNet} to estimate traffic speeds, and the evaluation set to assess the accuracy.

An input patch size $K_T= K_X = 64$ and output patch size $L_T= L_X = 32$ are used. Chosen patch sizes represent a balance between several factors: On the one hand, with larger $K_T,K_X$, an input patch contains more input data and can potentially relate measurements with each other. Thus, a potential correlation between two data points, that are distant in time and space, can be captured, which may increase the reconstruction accuracy. On the other hand, larger input patch sizes make it more difficult for the network to learn dependencies, since more permutations are possible. Furthermore, more convolution operations are necessary to process the increased layer sizes, which worsens run-time. An input patch, given $K_T, K_X=64$ with a $\Delta T = \unit[60]{s}$ and $\Delta X = \unit[100]{m}$, covers a stretch of $\unit[6.4]{km}$ and rounded $\unit[107]{min}$. The output size $L_T,L_X$ is chosen as half the input size. This parameter is based on similar considerations: One the one hand, since a space-time domain is divided into a set of output patches, a larger output patch reduces the run-time for the reconstruction of the entire domain because fewer patches need to be computed. On the other hand, considering the cells that are positioned at the boundary of any output patch (e.g. $i$ or $j$ equals zero or $N_X-1=N_T-1=31$), there are only $(K_T,L_T)/2$ or rather $(K_X-L_X)/2$ cells between boundary of output and boundary of input cell. This means, that $(64-32)/2=16$ cells, which equal $\unit[1.6]{km}$ and rounded $\unit[27]{min}$, containing potential raw data can be considered for the estimation of these output patch boundary cells. Assuming we would use $K_T=L_T$, this boundary difference would equal zero, with the consequence that cells at the output boundary, would be estimated based on a reduced number of input cells compared to inner cells of the output patch. In practice, this results in reconstruction artifacts at the cells boundaries, which reduce the overall reconstruction accuracy. 

For training, the input data is bound to $\unit[3]{km/h} \le v \le \unit[130]{km/h}$ in order to make them comparable to outputs of the \ac{ASM} and \ac{PSM}. Only part of all patches are used for training: the ones that belong to scenarios that are used for the subsequent analyses remain unseen for the neural network. The \textit{Adam} optimizer is applied \cite{kingma2017adam} to adapt the neural network weights during the training process. Weights are initialized with the \textit{Glorot Uniform} initializer \cite{glorot2010understanding}. A batch size of 32 is used. As error the \ac{IMAE} is used (see section \ref{sec:metric}), which returns for a scalar for each input-output sample. The average \ac{IMAE} of a batch is used as error for back-propagation.

Training is run for 50 epochs using TensorFlow on a workstation with an Intel Xeon CPU, 64GB RAM and an Nvidia Quatro RTX 4000. On this machine, training takes around three hours to finish.

\subsection{Qualitative Results}
The rightmost column in Figure \ref{fig:layer_details} depicts some of the layer outputs of the fully trained network given one example input patch. It is interesting to see how the information of the input patch gets encoded in the first parts of the neural network, resulting in the $4\times4$ matrices in layer 35. The subsequent decoding structure and further convolution layers upsample and combine the information, which finally results in the depicted output matrix. Apparently, the neural network is able to reconstruct the moving jams that are visible in the raw \ac{FCD}. Please note that the output matrix with smaller size needs to be interpreted as being located in the center of the input patch (see Figure \ref{fig:patch}).


\begin{figure}[tbh]

  \includegraphics[width=.7\textwidth]{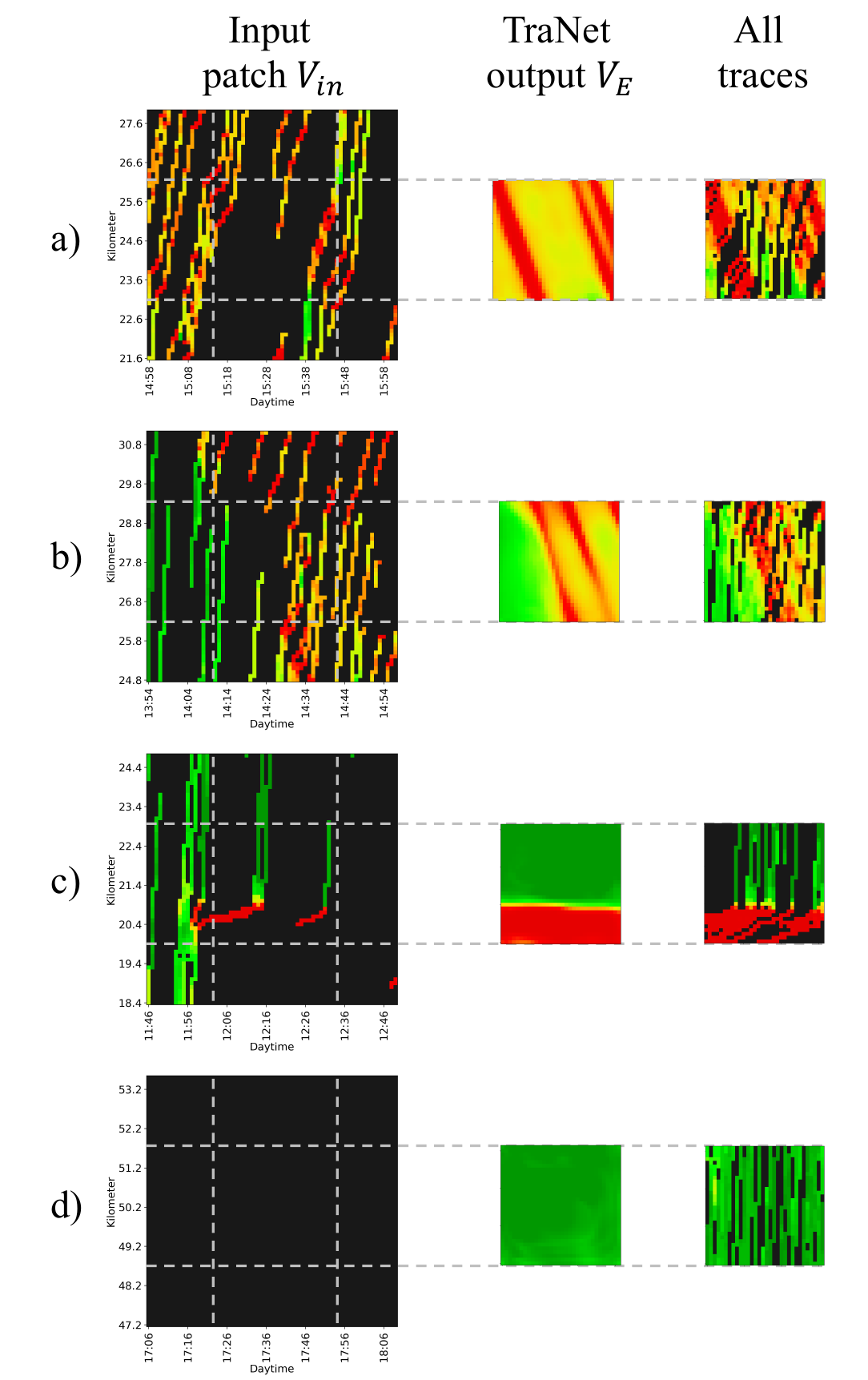}
\caption{Some examples of input patches, the respective computed output of \ac{TraNet} after training, applied on the input patch (unseen data), and all available data for comparison. (a) and (b): situations with moving congestion, (c): situation with stationary congestion and (d): no input is given}
\label{fig:examples_patches}
\end{figure}

\clearpage

Figure \ref{fig:examples_patches} displays some examples of input patches and respective output patches using trained \ac{TraNet}. In (a) and (b) data during shockwaves are displayed. \ac{TraNet} accomplishes to reconstruct these scenarios given sparse and noisy input data. Also the stationary congestion patch, displayed in (c) is reconstructed accurately. (d) displays the output if no data is given. In this case, \ac{TraNet} has learned automatically from given training data, that most likely, traffic state is in free-flow conditions, and outputs traffic speeds close to the set maximum speed of $\unit[130]{km/h}$.

In the following, an example is given that illustrates how \ac{TraNet} reconstructs a complete congestion scenario based on sparse data. Therefore, \ac{FCD} collected during several congestion patterns on December 13th, 2019 in the northbound direction, are utilized (see Figure \ref{fig:vall_examples} (d)). This scenario is selected since it comprises different congestion patterns. Localized congestion with average vehicle speeds at approximately \unit[50]{km/h} with a relatively long time of existence, probably due to a capacity drop at the on-ramp \cite{geroliminis.capacitydrop,Saberi.2013,Kerner.2004,trafficOsciAndCapacityDrop}, can be observed at e.g. kilometer 34 and 44. Inside these stationary congestions, moving jams emerge and propagate upstream (compare to synchronized traffic and \acp{WMJ} in the three-phase traffic theory \cite{Kerner.2004} or \textit{oscillating congested traffic} and \textit{stop and go waves} according to a classification presented in \cite{Helbing.2009}). These moving jams are commonly observed traffic patterns. Furthermore, a mega-jam (\cite{Kerner.2004}, also called \textit{homogeneous congested traffic} \cite{Helbing.2009}), occurred with a downstream front located at approximately kilometer 32. This is a third pattern which forms upstream of a strong bottleneck such as lane closures due to an incident. 

\begin{figure}  [tbh]
  \includegraphics[width=\textwidth]{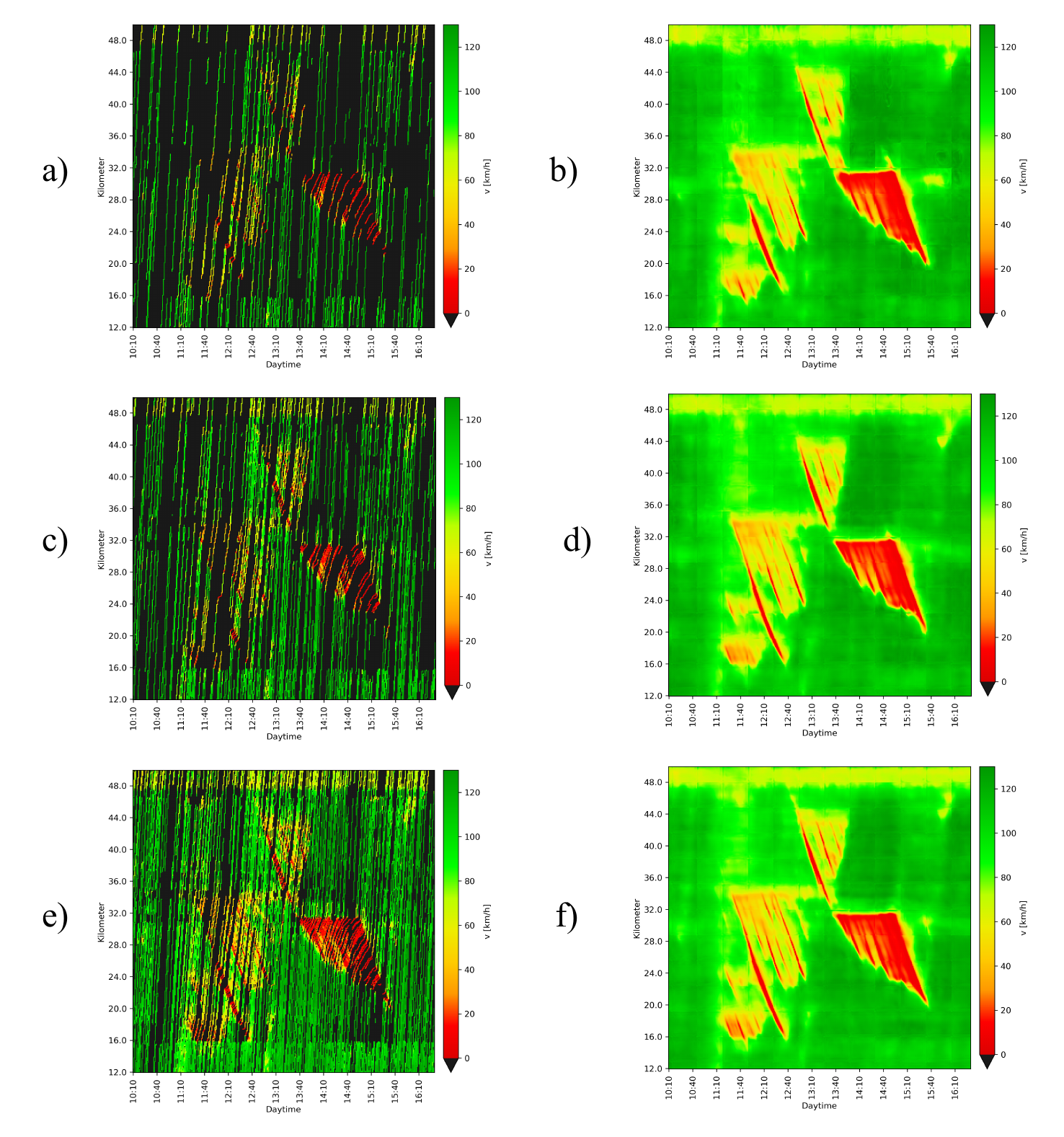}
  { \caption{ (a,c,e) input trajectory data to the network with $20\%, 50\%$ and $100\%$ of all available trajectories; (b,d,f) the estimated traffic speeds using \ac{TraNet}}
    \label{fig:tranet_applic} }
\end{figure}

Trajectory data collected during that scenario are randomly selected with a ratio of $p={0.2,0.5,1.0}$ (see first column in Figure \ref{fig:tranet_applic}). As mentioned in section \ref{sec:inout}, the domain is divided into overlapping input patches of size $K_T \times K_X$ such that the output patches of sizes $L_T \times L_X$ form a uniform grid. Boundaries that overlap the presented domain are set to undefined values in the input patches. For each input patch, an output is generated using \ac{TraNet}, and the resulting output patches are stitched together.

Figure \ref{fig:tranet_applic} displays the used input data and the respective output data. At first glance the estimated outputs for the three situations look very similar: the mega-jam starting at 1:40pm is reconstructed well in all three situations. Moreover, the several \acp{WMJ} that propagate upstream at different times ad locations within that domain are visible in all three examples. Nevertheless, there are also differences in the reconstruction output: for instance, the moving jam that emerges at approximately 11:30am at kilometer 34 which is visible clearly in (f) is estimated to originate in (b) significantly later, at 11:40am. Additionally, the fine sequence of moving jams is estimated more accurately in (f) than in (b). Another observation is that in (b), the grid boundaries of the output patches are visible, especially in free flow conditions. They are less visible with more data as in (f).

\begin{figure}[tbh]
  \includegraphics[width=\textwidth]{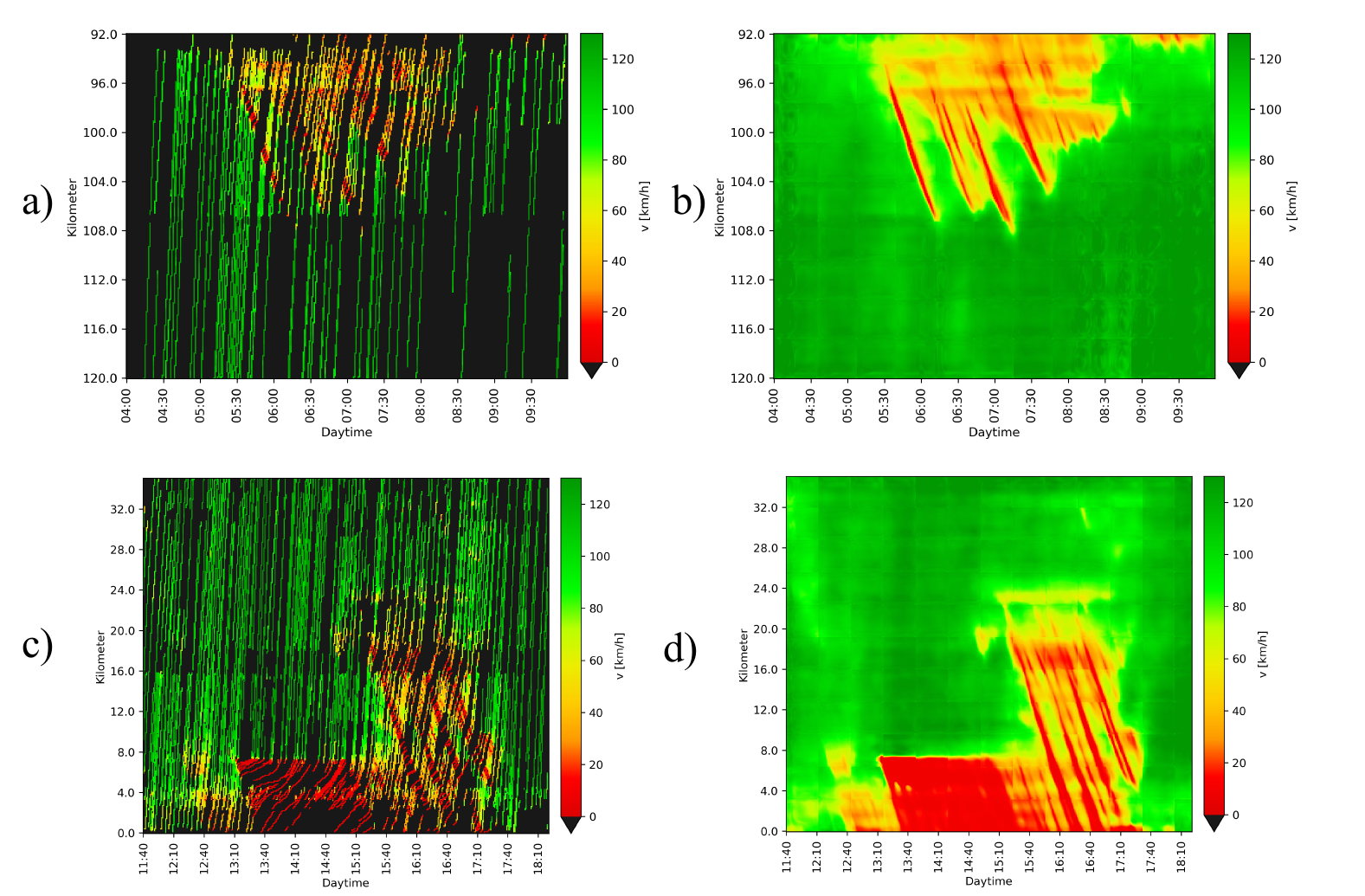} 
  { \caption{ (a,c) input trajectory data to the network as a sparse subset of trajectories based on the situation given in \ref{fig:vall_examples} a) and c); (b,d) the estimated traffic speeds using \ac{TraNet}}
    \label{fig:tranet_applic_2} }
\end{figure}

Figure \ref{fig:tranet_applic_2} visualizes two further examples of \ac{TraNet} applied to raw data of congestion scenarios visualized in Figure \ref{fig:vall_examples}. Also in these examples, \ac{TraNet} is able to reconstruct moving as well as stationary congestion with high accuracy based on the given data.

All in all, the outputs are quite promising as the neural network manages to reconstruct a situation even with little available data. Apparently it is able to differentiate between different patterns such as moving jams, localized traffic congestion with medium vehicle speeds as well as very low vehicle speeds upstream of a strong bottleneck. This is notable since the neural network was able to learn the reconstruction abilities given only a set of example data (excluding patches from this situation) and an error metric.

\subsection{Performance analysis} \label{sec:evaluation}
The following study shows the performance of \ac{TraNet} compared to other state-of-the-art approaches. As comparable approaches, the \ac{ASM}, the \ac{PSM}, and a naive smoothing method are presented. Since there do not exist published neural network approaches that tackle the problem, yet, and there are countless potential architectures of neural networks, which could be constructed, but which were outperformed by \ac{TraNet} in the experimental phase, no additional neural network is included in the evaluation. Rather, focus is on \ac{TraNet} as the best-performing approach out of all the architectures that were tried out.

\subsubsection{Isotropic Smoothing}
The naive smoothing method is a simple approach that smooths data using an isotropic kernel:
\begin{equation}
  V_{iso}(t,x) = \frac{1} { \sum_{(t_d,x_d,v)}^{V_{FCD}} \Phi(t-t_d,x-d_d)}  \sum_{(t_d,x_d,v)}^{V_{FCD}} \Phi(t-t_d,x-d_d) v 
\end{equation}
with 
\begin{equation}
  \Phi(t,x) = \exp \big( - \big| \frac{t}{\tau} \big| - \big| \frac{x}{\sigma} \big| \big)
\end{equation}
where $\tau$ and $\sigma$ are parameters to calibrate the gradient of the exponential decay of the kernel in time and space. They are set to $\tau=\unit[150]{s}$ and $\sigma=\unit[300]{m}$.

\subsubsection{\ac{ASM}}
\label{sec:ASM}
The \ac{ASM} is a well-known approach that is applied frequently in order to estimate traffic conditions retrospectively and in real-time \cite{Treiber.2002,Treiber.2011,Schreiter.2010,Kessler.2018.TRB,Rempe.2016,asm_rolling_smoothing}. Briefly summarized, raw data of a sparse input source are convolved in two traffic-characteristic directions: $v_{cong}$ denominating the wave speed in congested traffic conditions, and $v_{free}$ denominating the wave speed in free-flow conditions. The resulting matrices $V_{cong}(t,x)$ and $V_{free}(t,x)$ are combined cell-wise with an adaptive weighting:

  \begin{equation} \label{eq:ASM-v}
  V_{ASM}(t,x) = w(t,x)V_{cong}(t,x) + (1-w(t,x))V_{free}(t,x).
  \end{equation}
\noindent The weighting $w(t,x)$ is defined as:

  \begin{equation} \label{eq:ASM-w}
    w(t,x) = \frac{1}{2} \Big(  1+ \tanh \big( \frac{V_{thres} - \min (V_{cong}(t,x), V_{free}(t,x) )}{\Delta V}  \big) \Big)
  \end{equation}
\noindent with $V_{thres}$ a threshold where weight $w(t,x)$ equals $0.5$ and $\Delta V$ a parameter to control the steepness of the weight function. An analysis in \cite{Lint.2010} points out that travel times based on speeds reconstructed using the \ac{ASM} lack accuracy. Since the \ac{IMAE} also assesses inverse speeds, the proposed measure to smooth the inverse speeds is adopted here. Parameters are set according to \cite{Lint.2010}.

\subsubsection{PSM}
\label{sec:PSM}
The \ac{PSM} utilizes principles of the \ac{ASM} and extends its concepts based on the Three-Phase theory \cite{Rempe.2017.PSM}. The motivation for its development is that the \ac{ASM} applied to \ac{FCD} is not able to accurately reconstruct stationary congestion patterns \cite{Rempe.2017.PSM,Treiber.2011}. 

In order to improve on that aspect, the \ac{PSM} applies a three-step approach. For the details of the method, the reader is referred to the original paper \cite{Rempe.2017.PSM}. First, raw data is processed using specific convolution kernels and several empirically motivated criteria in order to determine space-time phase regions. For instance, synchronized congestion pattern tend to be localized at a bottleneck such as an on-ramp. \acp{WMJ} propagate upstream and maintain a relatively constant downstream phase front speed. In the second step, the extracted phase regions are used to smooth data within their corresponding traffic phase. Third, phases probabilities and speeds are aggregated into a final speed output.

Since the method is specialized for the reconstruction of traffic speeds from sparse \ac{FCD}, it is included in the evaluation. Parameters are taken according to the original paper.

\subsubsection{CNN network}
An approach recently published \cite{deepCNNtrafficdata} applies a \ac{CNN} with an encoding-decoding structure to trajectory data in order to impute traffic speeds. The network comprises three 2-D convolution layers with kernel sizes $(5,5)$ with a ReLU activation, followed by 2-D MaxPooling layers with stride $(2,2)$. The number of filters increases with decreasing domain size: $(8,32,64)$. The decoding sequence of the network comprises three \ac{CNN} layer with ReLU activation and filter sizes $(3,3)$ and $(5,5)$ as well as a decreasing number of filters: $(64,32,8,3)$. As upsampling layer a 2-D nearest neighbor layer is applied. 

We tried to rebuild the model as accurately as possible, but needed to do minor modifications to make the approaches comparable: First, the authors of the original paper process RGB data with the goal to allow the network to distinguish between data points and missing data. We input the data similarly to \ac{TraNet} as two layers: normalized speeds and grid occupancies. This facilitates learning for the network. Second, since in the input patches are larger than the output patches in order to avoid reconstruction artifacts at patch boundaries (see section \ref{sec:trafnet}), this \ac{CNN} requires one upsampling layer less in order to scale up to the requested output size.

We train this network with the same data as \ac{TraNet} and stop after training for 50 epochs as well. At this point the validation error starts to increase again, which is an indication for a network to start overfitting (see \cite{Bishop.2013}). In the following, the network is denotes as $CNN6$ referring to the 6 \ac{CNN} layers.

\subsubsection{Results}
\begin{figure}
  \includegraphics[width=\textwidth]{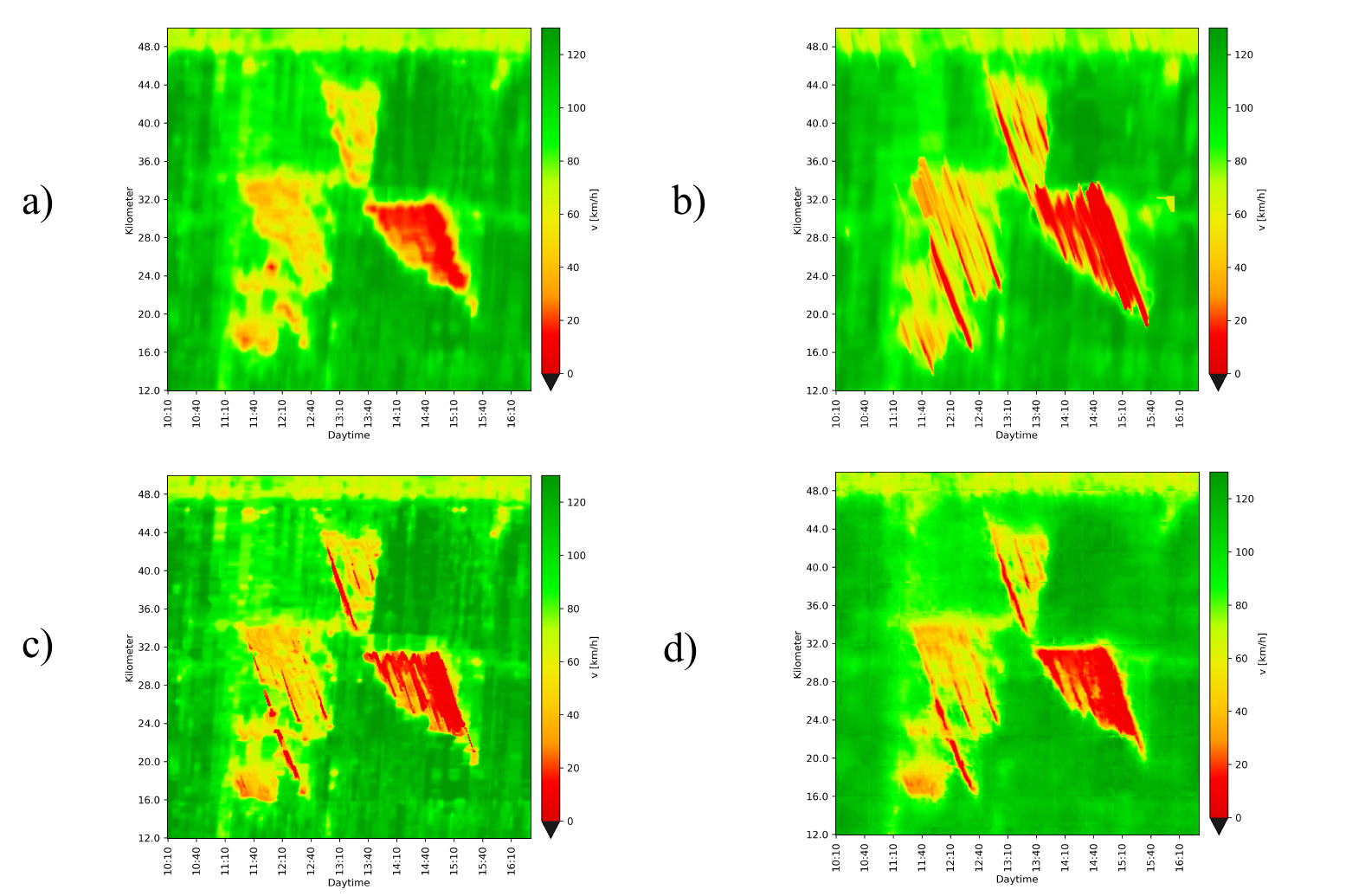}    
   \caption{ The estimated speeds using the (a) isotropic method, (b) the \ac{ASM}, (c) the \ac{PSM} and the (d) $CNN6$. As input raw data visualized in Figure \ref{fig:tranet_applic} (c) using 50\% of all available data is considered. }
    \label{fig:recons_all} 
\end{figure}

Figure \ref{fig:recons_all} displays the reconstructed speeds calculated by the isotropic smoothing, the \ac{ASM} and the \ac{PSM} using the raw trajectory data with a ratio of $p=0.5$ as visualized in Figure \ref{fig:tranet_applic} (c). The isotropic smoothing results in a relatively blurry traffic speed representation: the transitions between free and congested traffic, especially the mega-jam between 1pm and 3pm upstream of kilometer 32, are significantly smoother than the transitions visible in raw data (compare Figure \ref{fig:tranet_applic} (e)). Furthermore, the isotropic smoothing fails to reconstruct the moving jams. The \ac{ASM}, using its two smoothing filters, reconstructs these much more accurately. Nevertheless, it overestimates the extent of the mega-jam: the congestion front is 'spills over' the downstream jam front. The \ac{PSM} reduces these issues and is also able to reconstruct the moving jams. However, when applied to sparse data, there appear some reconstruction artifacts: not all moving jams are reconstructed correctly, and the downstream jam front of the mega-jam is 'ragged'.

\begin{figure}[tbh]
  \includegraphics[width=\textwidth]{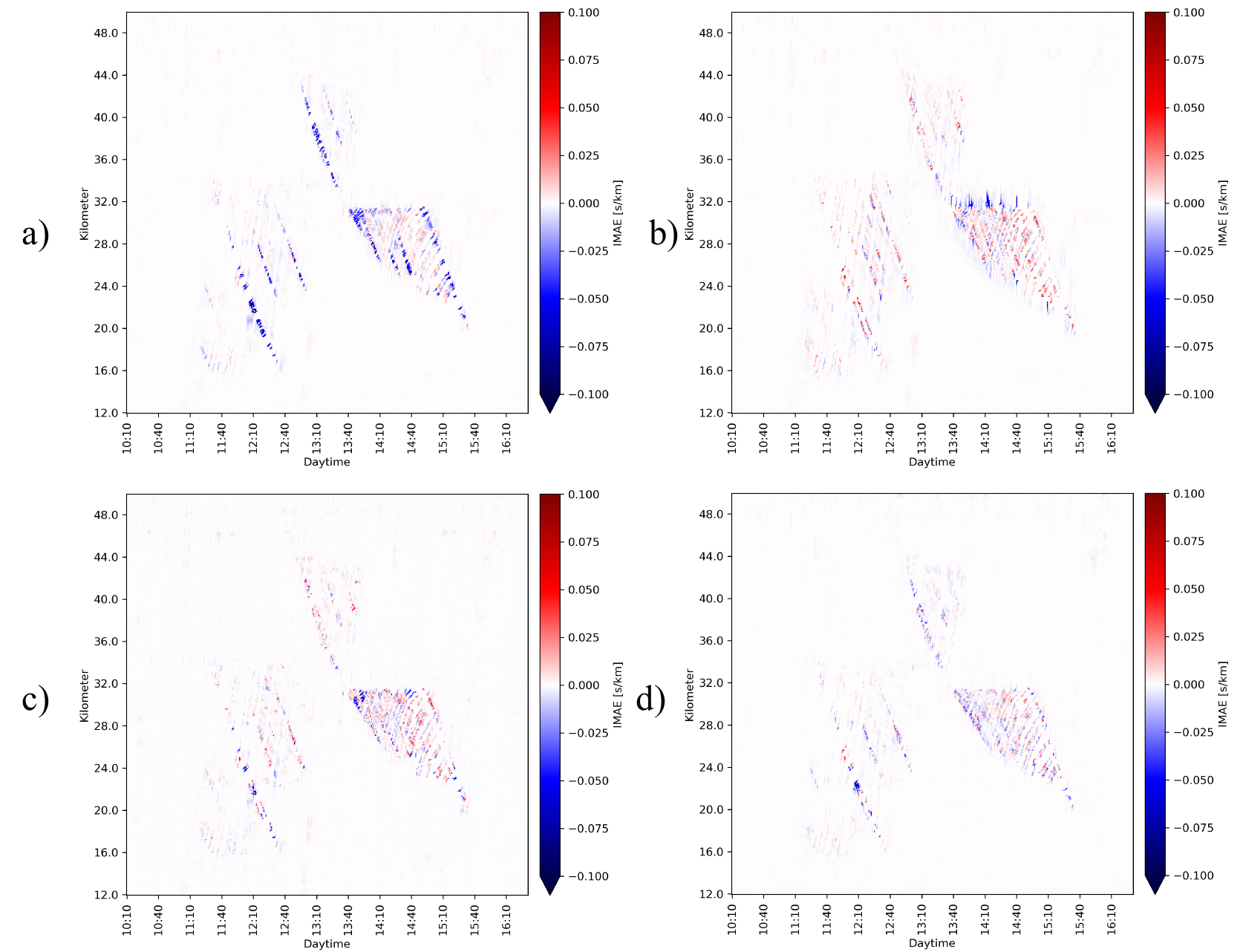}  
  \caption{Error differences of the (a) isotropic smoothing, the (b) \ac{ASM}, the (c) \ac{PSM} and the (d) $CNN6$ compared to \ac{TraNet}. Blue colors indicate regions where \ac{TraNet} reconstructed the speed more accurately than the other approach; red colors indicate that the compared approach returns more accurate speeds.}\label{fig:diffplot} 
\end{figure}

In order to compare estimated speeds more distinctively, we consider the error difference $IMAE_\Delta$ of two reconstructed speeds $V_1$ and $V_2$, compared to the \ac{GT} $V_{GT}$, valid for all grid cells $(t,x) \in V_{GT}$ containing a valid speed:
\begin{equation}
  IMAE_\Delta(V_1,V_2) = \big|\frac{1}{V_1(t,x)} - \frac{1}{V_{GT}(t,x)}\big| - \big|\frac{1}{V_2(t,x)} - \frac{1}{V_{GT}(t,x)}\big|
\end{equation}
Thus, $IMAE_\Delta$ compares the estimation errors of two approaches: If the error of algorithm 1 is larger, the resulting output is positive, if they are equal it is zero, and if the error of algorithm 2 is larger, the outcome is negative. Figure \ref{fig:diffplot} plots the error differences of the isotropic smoothing (a), the \ac{ASM} (b) and the \ac{PSM} (c) compared to \ac{TraNet}. As $V_{GT}$, all data samples are utilized. The output of \ac{TraNet} is taken as speed matrix $V_2$. Consequently, blue colors depict cells where \ac{TraNet} reconstructed the speed more accurately than the other approach, and red color the cells with lower reconstruction accuracy.

These plots emphasize the results of the aforementioned observations in a direct comparison: \ac{TraNet}, with significantly larger blue regions reconstructs most relevant cells more accurately than the isotropic smoothing (a). With respect to the \ac{ASM}, the difference is smaller and more red regions are visible. Still, the overestimation of the congested regions of the stationary congestion are depicted with dark blue, indicating a larger error of the \ac{ASM}. The comparison with the \ac{PSM} shows a mixed result. Overall, the blue color dominates, though, in some situations the \ac{PSM} provides higher accuracies. For instance, the \ac{WMJ} originating at kilometer 44 at 12:30pm is captured better using the \ac{PSM}. At the boundaries of the mega-jam, the dark blue colors indicate the space-time regions where \ac{TraNet} outperforms the \ac{PSM} significantly. The error plot of \ac{TraNet} compared to CNN6 shows that \ac{TraNet} estimates speeds during the \ac{WMJ} better and tends to estimates also better during the stationary congestion. This result indicates, that the network architecture of the CNN6 is not able to associate low traffic speeds of spatio-temporally distant traces sufficiently well in order to reconstruct the moving and stationary jams accurately.

\begin{figure}[tbh]
	\centering
	\includegraphics[width=.7\textwidth]{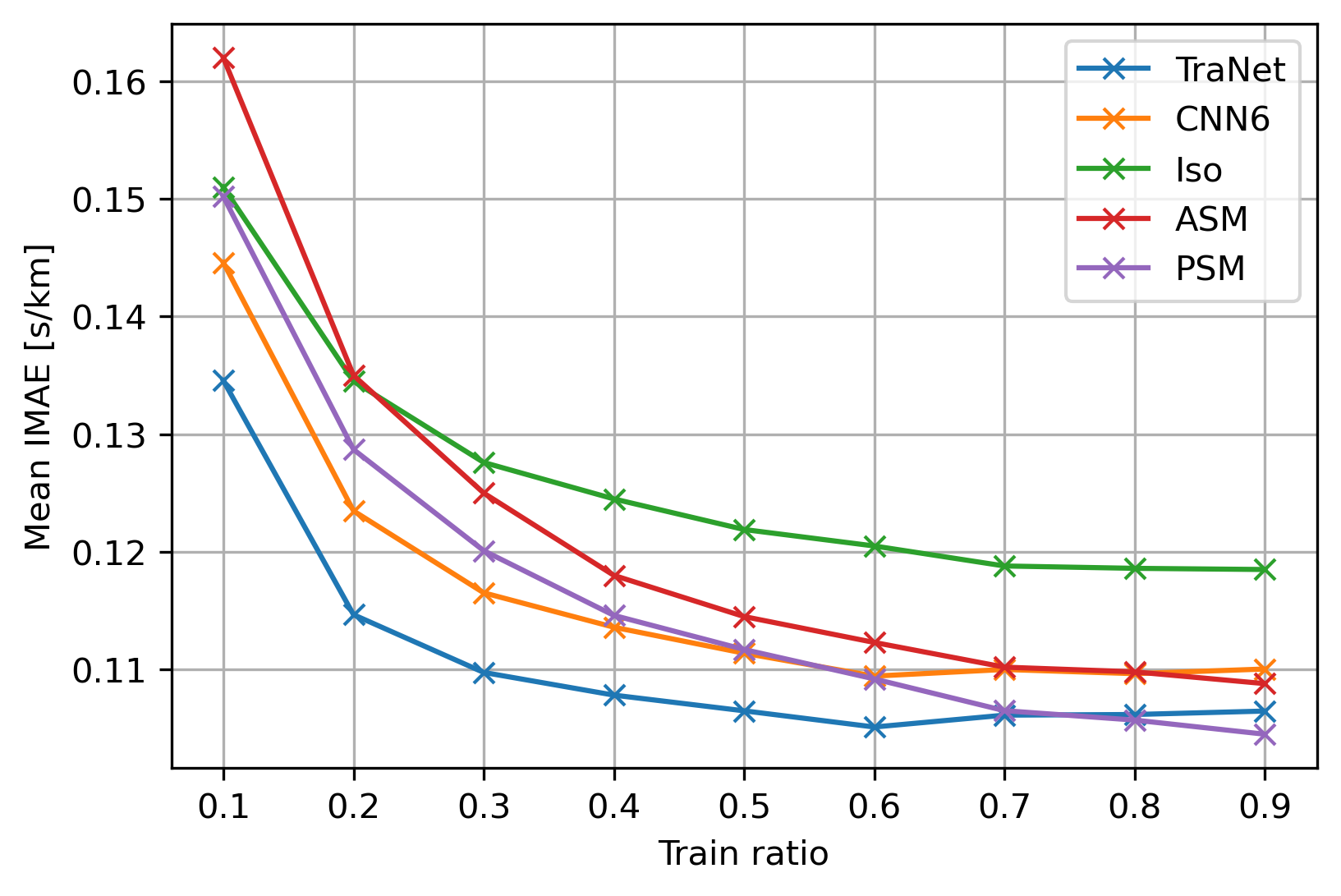}
	\caption{Reconstruction accuracy considering the IMAE of the four methods with respect to a varying data density (i.e. train ratio).}
	\label{fig:imae_comparison}
\end{figure}

After a rather qualitative analysis, a quantitative comparison gives further insights into the reconstruction accuracies. Since sparse data is a common issue, and it is difficult to foresee how much data will be available, a generally applicable reconstruction method is able to return accurate reconstructions for problems with both, sparse and dense data. Therefore, in the following it is studied how well the algorithms perform under varying input data density. Given the same scenario, the ratio of train and test trajectories $p$ is varied in the range of $[0.1-0.9]$. Again, train data is used to reconstruct the situation. The remaining test data is used to assess its reconstruction quality. Over 100 iterations, a random split between training and test data is done according to the ratio $p$, and the $IMAE$ of the reconstructed speeds is calculated using the test traces. The resulting average \acp{IMAE} with respect to the training set and the utilized algorithm are depicted in Figure \ref{fig:imae_comparison}.

As is visible, all algorithms achieve a lower error, i.e. higher accuracy with higher data density. \ac{TraNet} returns the most accurate results up to $p=0.8$. Especially compared to the isotropic smoothing the error is significantly lower. The \ac{PSM} outperforms the \ac{ASM} for all data densities, but is still significantly less accurate than \ac{TraNet} at lower densities. Given high data densities, \ac{TraNet} and \ac{PSM} provide comparable results. The CNN6 approach provides more accurate estimates for low data densities than \ac{ASM} and \ac{PSM}, with high densities the \ac{PSM} gives slightly better results. Thus, we can confirm the superiority of CNN6 over the ASM, which is a key result of the study done on the NGSIM dataset in \cite{deepCNNtrafficdata}, using our dataset. Still, \ac{TraNet} achieves more accurate results than CNN6 for all data densities. This shows, that the more significantly deeper structure of \ac{TraNet} with its feed-forward paths has better learning and estimation capabilities than the simpler CNN6.

To summarize the qualitative and quantitative comparison with state-of-the-art approaches, the novel \ac{TraNet} is able to reconstruct traffic speeds based on sparse data with higher accuracy. Especially in situations with low data density, the gain compared to the other approaches is significant. 

\section{Discussion}
\label{sec:discussion}
The approach to apply a \ac{DNN} to the problem of traffic speed reconstruction with sparse data resulted in several findings:
\begin{enumerate}
  \item Given sparse, grid-based speed data as input, the neural network can be trained using only input-output samples and an error metric, such that it is ultimately able to produce a complete speed output.
  \item The neural network is able to reconstruct several empirical congestion patterns distinctively (see \cite{Helbing.,Kerner.2004}), e.g. (i) stationary (localized) congestion at bottlenecks with spontaneously emerging (narrow) moving jams, (ii) \acp{WMJ} which propagate upstream for kilometers and low vehicle speeds, and (iii) severe stationary congestion at strong bottlenecks (mega-jams), where vehicles tend to queue in order to pass the bottleneck location.
  \item With respect to the \ac{IMAE} metric, the average reconstructed speeds are more accurate than results of comparable approaches. The gain in accuracy is significant if low to medium data densities are available as input. In comparison to simpler \ac{CNN}, it appears that the increased network complexity of \ac{TraNet} allows to achieve higher results.
\end{enumerate}

The results are quite encouraging as they show that, given only raw data and a metric, the deep learning approach is able to extract relevant patterns from data and apply them to new, unseen data. Thus, the approach in a way automates the process of traffic modeling: it seeks to find patterns in data and generalize the problem by processing congestion scenarios while evaluating the success of (many) numerical experiments. Although the method has been trained and evaluated solely with \ac{FCD} in this study, due the generality of the approach, it is expected that all types of grid-based speeds may serve as data source.

Despite the higher accuracy and automated learning process, there are some specifics in the application of \acp{DNN} that need to be considered when applying them:
\begin{enumerate}
  \item The model can only be as good as the given data. This means, if the trained network is applied to a different data source, e.g. speeds collected by loops and the neural network is not re-trained, it is likely that the output results will lack accuracy. In this case, one should have a look at a technique called \textit{transfer learning} \cite{goodfellow2016deep}, the idea of which is to fix most of the neural network's parameters and re-train the network with input-output samples of the problem it is supposed to adapt to.
  \item The computations done within a \ac{DNN} are complex. Therefore, it is difficult to estimate the output of a neural network in advance (black box). 
  \item If the fundamental problem, that the \ac{DNN} is supposed to solve, changes (e.g. the desired grid cell size), it usually needs to be re-trained, whereas a method such as the \ac{ASM} can be adapted using one of its configuration parameters. 
\end{enumerate}

\section{Conclusion} \label{sec:conclusion}

This paper approaches the problem of traffic speed estimation using probe speeds. An adapted \ac{DNN} architecture is developed that requires for training and application only sparse speed data given within a space-time domain of variable size. The domain is decomposed into regularly sized space-time matrices containing the data itself, and an occupancy matrix indicating the existence of measurements. Using a large set of probe data collected during 42 congestion scenarios on a German motorway, the network is trained. Training data is augmented using a random sampling process in order to make the network's performance invariant to shifted data in time and space, and invariant to the prevailing data density. During training, the network learns to reconstruct various congestion patterns such as moving and stationary congestion.

The fully trained network is applied to data of a scenario that is unseen to the network. The reconstructed speeds with respect to the data density are studied qualitatively, and compared to the estimates computed using an isotropic smoothing, the \ac{ASM}, the \ac{PSM} and a standard \ac{CNN}. The resulting error plots reveal that the \ac{DNN} reconstructs certain traffic features such as moving and stationary congestion more accurately than the methods in comparison. In a quantitative comparison of the different approaches, the \ac{DNN} outperforms the other approaches for a variety of input data densities with respect to the \ac{IMAE}, significantly.

The result show that a deep neural network, that does not contain prior information about traffic dynamics is able to learn solely from raw data, such that it is capable of reproducing accurate congestion patterns given little data. The higher accuracy of the trained \ac{DNN} with its more complex architecture compared to a simple convolutional encoding-decoding network shows, that this problem class benefits from a more dedicated network architecture. The advantages of the approach, namely, that only data of single sparse data source is required for training, the \ac{DNN} can be applied to variable space-time domains and its estimation accuracy is high when applied to data of an unseen congestion pattern, contribute to the general applicability of this approach.

Further fields of research that may build upon this \ac{DNN}, are approaches that may require less data to learn, or reconstruct speeds even more accurately. Potential candidates include \acp{PINN} or Generative Adversarial Imputation Nets \cite{gain_traffic}.

\hspace{10cm}

\section*{Acronyms}
\begin{acronym}[GASM]
	\acro{ASM}{Adaptive Smoothing Method}
	\acro{PSM}{Phase-Based Smoothing Method}
	\acro{FCD}{Floating-Car Data}
 	\acro{IMAE}{Inverse Mean Absolute Error}
  \acro{WMJ}{Wide Moving Jam}
  \acro{CNN} {Convolutional Neural Network}
  \acro{DNN} {Deep Neural Network}
  \acro{FD} {Fundamental Diagram}
  \acro{LWR} {Lighthill-Whitham-Richards}
  \acro{GPS} {Global Positioning System}
  \acro{PDE} {Partial Differential Equation}
  \acro{PINN} {Physics-Informed Neural Network}
  \acro{SVR} {Support Vector Regression}
  \acro{LSTM} {Long-Short Term Memory}
  \acro{TraNet} {Traffic Network}
  \acro{GT} {Ground Truth}
  \acro{ReLU} {Rectified Linear Unit}
\end{acronym}

\hspace{10cm}
\section*{Information on Funding}
This research did not receive any specific grant from funding agencies in the public, commercial, or not-for-profit sectors.

\hspace{10cm}
\section*{Acknowledgment}
Special thanks to Allister Loder and Patrick Malcolm for the valuable inputs und proof-reading.

\newpage
\bibliographystyle{trb}
\bibliography{trb_template}

\begin{thebibliography}{76}
\providecommand{\natexlab}[1]{#1}

\bibitem[{{van Lint} et~al.(2005){van Lint}, Hoogendoorn, and {van
  Zuylen}}]{vanLint.2005}
{van Lint}, J., S.~P. Hoogendoorn, and H.~J. {van Zuylen}, {Accurate freeway
  travel time prediction with state-space neural networks under missing data}.
  \emph{{Transportation Research Part C: Emerging Technologies}}, Vol.~13, No.
  5-6, 2005, pp. 347--369.

\bibitem[{Ni and Wang(2008)}]{trajectoryRecons.2008}
Ni, D. and H.~Wang, Trajectory Reconstruction for Travel Time Estimation.
  \emph{Journal of Intelligent Transportation Systems}, Vol.~12, No.~3, 2008,
  pp. 113--125.

\bibitem[{{Hou} et~al.(2007){Hou}, {Xu}, and
  {Zhong}}]{freewaycontrolspeed.2007}
{Hou}, Z., J.~{Xu}, and H.~{Zhong}, Freeway Traffic Control Using Iterative
  Learning Control-Based Ramp Metering and Speed Signaling. \emph{IEEE
  Transactions on Vehicular Technology}, Vol.~56, No.~2, 2007, pp. 466--477.

\bibitem[{Han et~al.(2017)Han, Hegyi, Yuan, Hoogendoorn, Papageorgiou, and
  Roncoli}]{HAN2017405}
Han, Y., A.~Hegyi, Y.~Yuan, S.~Hoogendoorn, M.~Papageorgiou, and C.~Roncoli,
  Resolving freeway jam waves by discrete first-order model-based predictive
  control of variable speed limits. \emph{Transportation Research Part C:
  Emerging Technologies}, Vol.~77, 2017, pp. 405 -- 420.

\bibitem[{Chen and Chien(2001)}]{freewayTravelTimeprediction.2001}
Chen, M. and S.~I.~J. Chien, Dynamic Freeway Travel-Time Prediction with Probe
  Vehicle Data: Link Based Versus Path Based. \emph{Transportation Research
  Record}, Vol. 1768, No.~1, 2001, pp. 157--161.

\bibitem[{Coifman and Kim(2009)}]{COIFMAN2009349}
Coifman, B. and S.~Kim, Speed estimation and length based vehicle
  classification from freeway single-loop detectors. \emph{Transportation
  Research Part C: Emerging Technologies}, Vol.~17, No.~4, 2009, pp. 349 --
  364.

\bibitem[{Herrera et~al.(2010)Herrera, Work, Herring, Ban, Jacobson, and
  Bayen}]{Herrera.2010b}
Herrera, J.~C., D.~B. Work, R.~Herring, X.~Ban, Q.~Jacobson, and A.~M. Bayen,
  {Evaluation of traffic data obtained via GPS-enabled mobile phones: The
  Mobile Century field experiment}. \emph{{Transportation Research Part C:
  Emerging Technologies}}, Vol.~18, No.~4, 2010, pp. 568--583.

\bibitem[{Lighthill and Whitham(1955)}]{Lighthill.1955}
Lighthill, M. and G.~Whitham, {On kinematic waves II. A theory of traffic flow
  on long crowded roads.} \emph{{Proceedings of the Royal Society}}, Vol. A
  229, 1955, pp. 317--345.

\bibitem[{Richards(1956)}]{Richards.1956}
Richards, P., {Shock waves on the highway}. \emph{{Operations Research}},
  Vol.~4, 1956, pp. 42--51.

\bibitem[{Herrera and Bayen(2010)}]{Herrera.2010}
Herrera, J.~C. and A.~M. Bayen, {Incorporation of Lagrangian measurements in
  freeway traffic state estimation}. \emph{{Transportation Research Part B:
  Methodological}}, Vol.~44, No.~4, 2010, pp. 460--481.

\bibitem[{Treiber and Helbing(2003)}]{Treiber.2003ASM}
Treiber, M. and D.~Helbing, {An adaptive smoothing method for traffic state
  identification from incomplete information}. \emph{{Interface and Transport
  Dynamics}}, Vol.~32, 2003, pp. 343--360.

\bibitem[{Van~Lint and Hoogendoorn(2010)}]{gasm2010}
Van~Lint, J. and S.~P. Hoogendoorn, A Robust and Efficient Method for Fusing
  Heterogeneous Data from Traffic Sensors on Freeways. \emph{Computer-Aided
  Civil and Infrastructure Engineering}, Vol.~25, No.~8, 2010, pp. 596--612.

\bibitem[{Rempe et~al.(2017{\natexlab{a}})Rempe, Franeck, Fastenrath, and
  Bogenberger}]{Rempe.2017.PSM}
Rempe, F., P.~Franeck, U.~Fastenrath, and K.~Bogenberger, A phase-based
  smoothing method for accurate traffic speed estimation with floating car
  data. \emph{Transportation Research Part C: Emerging Technologies}, Vol.~85,
  2017{\natexlab{a}}, pp. 644 -- 663.

\bibitem[{Palmer et~al.(2011)Palmer, Rehborn, and Kerner}]{palmer2011asda}
Palmer, J., H.~Rehborn, and B.~Kerner, ASDA and FOTO Models based on Probe
  Vehicle Data. \emph{Traffic Engineering \& Control, Hemming}, Vol.~4, 2011,
  pp. 183--191.

\bibitem[{Work et~al.(2010)Work, Blandin, Tossavainen, Piccoli, and
  Bayen}]{Work.2010}
Work, D.~B., S.~Blandin, O.~P. Tossavainen, B.~Piccoli, and A.~M. Bayen, {A
  Traffic Model for Velocity Data Assimilation}. \emph{{Applied Mathematics
  Research eXpress}}, 2010.

\bibitem[{Bekiaris-Liberis et~al.(2016)Bekiaris-Liberis, Roncoli, and
  Papageorgiou}]{BekiarisLiberis.2016}
Bekiaris-Liberis, N., C.~Roncoli, and M.~Papageorgiou, {Highway Traffic State
  Estimation With Mixed Connected and Conventional Vehicles}. \emph{{IEEE
  Transactions on Intelligent Transportation Systems}}, 2016, pp. 1--14.

\bibitem[{Raissi et~al.(2017)Raissi, Perdikaris, and
  Karniadakis}]{raissi2017physics}
Raissi, M., P.~Perdikaris, and G.~E. Karniadakis, Physics Informed Deep
  Learning (Part I): Data-driven Solutions of Nonlinear Partial Differential
  Equations, 2017.

\bibitem[{Yuan et~al.(2021)Yuan, Zhang, Yang, and Zhe}]{YUAN202188}
Yuan, Y., Z.~Zhang, X.~T. Yang, and S.~Zhe, Macroscopic traffic flow modeling
  with physics regularized Gaussian process: A new insight into machine
  learning applications in transportation. \emph{Transportation Research Part
  B: Methodological}, Vol. 146, 2021, pp. 88--110.

\bibitem[{He et~al.(2016)He, Zhang, Ren, and Sun}]{resnet}
He, K., X.~Zhang, S.~Ren, and J.~Sun, Deep residual learning for image
  recognition. In \emph{Proceedings of the IEEE conference on computer vision
  and pattern recognition}, 2016, pp. 770--778.

\bibitem[{LeCun et~al.(2015)LeCun, Bengio, and Hinton}]{LeCun2015}
LeCun, Y., Y.~Bengio, and G.~Hinton, Deep learning. \emph{Nature}, Vol. 521,
  No. 7553, 2015, pp. 436--444.

\bibitem[{Ma et~al.(2017)Ma, Dai, He, Ma, Wang, and Wang}]{cnnprediction}
Ma, X., Z.~Dai, Z.~He, J.~Ma, Y.~Wang, and Y.~Wang, Learning Traffic as Images:
  A Deep Convolutional Neural Network for Large-Scale Transportation Network
  Speed Prediction. \emph{Sensors}, Vol.~17, 2017, p. 818.

\bibitem[{{Lana} et~al.(2018){Lana}, {Del Ser}, {Velez}, and
  {Vlahogianni}}]{lana.prediction}
{Lana}, I., J.~{Del Ser}, M.~{Velez}, and E.~I. {Vlahogianni}, Road Traffic
  Forecasting: Recent Advances and New Challenges. \emph{IEEE Intelligent
  Transportation Systems Magazine}, Vol.~10, No.~2, 2018, pp. 93--109.

\bibitem[{Vlahogianni and Karlaftis(2013)}]{Vlahogianni.2013}
Vlahogianni, E. and M.~Karlaftis, {Testing and Comparing Neural Network and
  Statistical Approaches for Predicting Transportation Time Series}.
  \emph{{Transportation Research Record: Journal of the Transportation Research
  Board}}, Vol. 2399, 2013, pp. 9--22.

\bibitem[{{Vinayakumar} et~al.(2017){Vinayakumar}, {Soman}, and
  {Poornachandran}}]{deepprediction}
{Vinayakumar}, R., K.~P. {Soman}, and P.~{Poornachandran}, Applying deep
  learning approaches for network traffic prediction. In \emph{2017
  International Conference on Advances in Computing, Communications and
  Informatics (ICACCI)}, 2017, pp. 2353--2358.

\bibitem[{Dargan et~al.(2019)Dargan, Kumar, Ayyagari, and
  Kumar}]{Dargan2019ASO}
Dargan, S., M.~Kumar, M.~R. Ayyagari, and G.~Kumar, A Survey of Deep Learning
  and Its Applications: A New Paradigm to Machine Learning. \emph{Archives of
  Computational Methods in Engineering}, 2019, pp. 1--22.

\bibitem[{Ronneberger et~al.(2015)Ronneberger, Fischer, and
  Brox}]{ronneberger2015unet}
Ronneberger, O., P.~Fischer, and T.~Brox, U-Net: Convolutional Networks for
  Biomedical Image Segmentation. In \emph{Medical Image Computing and
  Computer-Assisted Intervention -- MICCAI 2015} (N.~Navab, J.~Hornegger, W.~M.
  Wells, and A.~F. Frangi, eds.), Springer International Publishing, Cham,
  2015, pp. 234--241.

\bibitem[{{van Lint} and Djukic(2014)}]{vanLint.2014}
{van Lint}, H. and T.~Djukic, {Applications of Kalman Filtering in Traffic
  Management and Control}. \emph{{New Directions in Informatics, Optimization,
  Logistics, and Production}}, 2014, pp. 59--91.

\bibitem[{Suzuki et~al.(2003)Suzuki, Nakatsuji, and Nanthawichit}]{Suzuki.2003}
Suzuki, H., T.~Nakatsuji, and C.~Nanthawichit, {Application of Probe Vehicle
  Data for Real-Time Traffic State Estimation and Short-Term Travel Time
  Prediction on a Freeway}. \emph{{Transportation Research Record: Journal of
  the Transportation Research Board}}, Vol. 1855, 2003, pp. 49--59.

\bibitem[{Aw and Rascle(2000)}]{Aw.2000}
Aw, A. and M.~Rascle, {Resurrection of Second Order Models of Traffic Flow}.
  \emph{{SIAM Journal on Applied Mathematics}}, Vol.~60, No.~3, 2000, pp.
  916--938.

\bibitem[{Wang and Papageorgiou(2005)}]{Wang.2005}
Wang, Y. and M.~Papageorgiou, {Real-time freeway traffic state estimation based
  on extended Kalman filter: a general approach}. \emph{{Transportation
  Research Part B: Methodological}}, Vol.~39, No.~2, 2005, pp. 141--167.

\bibitem[{Work et~al.(2008)Work, Tossavainen, Blandin, Bayen, Iwuchukwu, and
  Tracton}]{Work.}
Work, D.~B., O.-P. Tossavainen, S.~Blandin, A.~M. Bayen, T.~Iwuchukwu, and
  K.~Tracton, {An ensemble Kalman filtering approach to highway traffic
  estimation using GPS enabled mobile devices}. In \emph{{2008 47th IEEE
  Conference on Decision and Control}}, 2008, pp. 5062--5068.

\bibitem[{Work et~al.(2009)Work, Tossavainen, Jacobson, and Bayen}]{Work.b}
Work, D.~B., O.-P. Tossavainen, Q.~Jacobson, and A.~M. Bayen, {Lagrangian
  sensing: traffic estimation with mobile devices}. In \emph{{2009 American
  Control Conference}}, 2009, pp. 1536--1543.

\bibitem[{Treiber and Kesting(2013)}]{Treiber.2013}
Treiber, M. and A.~Kesting, \emph{{Traffic Flow Dynamics}}. {Springer Berlin
  Heidelberg}, Berlin, Heidelberg, 2013.

\bibitem[{Kerner(2004)}]{Kerner.2004}
Kerner, B.~S., \emph{{The Physics of Traffic}}. Springer, New York, 2004.

\bibitem[{{van Lint} and Hoogendoorn(2009)}]{vanLint.2009}
{van Lint}, H. and S.~Hoogendoorn, {A Robust and Efficient Method for Fusing
  Heterogeneous Data from Traffic Sensors on Freeways}. \emph{{Computer-Aided
  Civil and Infrastructure Engineering}}, Vol.~24, 2009, pp. 1--17.

\bibitem[{Rempe et~al.(2017{\natexlab{b}})Rempe, Kessler, and
  Bogenberger}]{Rempe.MITITS}
Rempe, F., L.~Kessler, and K.~Bogenberger, Fusing probe speed and flow data for
  robust short-term congestion front forecasts. In \emph{2017 5th IEEE
  International Conference on Models and Technologies for Intelligent
  Transportation Systems (MT-ITS)}, 2017{\natexlab{b}}, pp. 31--36.

\bibitem[{Rempe et~al.(2016{\natexlab{a}})Rempe, Franeck, Fastenrath, and
  Bogenberger}]{Rempe.2016GASM}
Rempe, F., P.~Franeck, U.~Fastenrath, and K.~Bogenberger, Online freeway
  traffic estimation with real floating car data. In \emph{2016 IEEE 19th
  International Conference on Intelligent Transportation Systems (ITSC)}, IEEE,
  2016{\natexlab{a}}, pp. 1838--1843.

\bibitem[{Kerner(1999)}]{Kerner.1999}
Kerner, B.~S., The physics of traffic. \emph{Physics World}, Vol.~12, No.~8,
  1999, pp. 25--30.

\bibitem[{Kerner et~al.(2004)Kerner, Rehborn, Aleksic, and Haug}]{Kerner.2004b}
Kerner, B.~S., H.~Rehborn, M.~Aleksic, and A.~Haug, {Recognition and tracking
  of spatial--temporal congested traffic patterns on freeways}.
  \emph{{Transportation Research Part C: Emerging Technologies}}, Vol.~12,
  No.~5, 2004, pp. 369--400.

\bibitem[{Rempe et~al.(2017{\natexlab{c}})Rempe, Franeck, Fastenrath, and
  Bogenberger}]{rempe2017phase}
Rempe, F., P.~Franeck, U.~Fastenrath, and K.~Bogenberger, A phase-based
  smoothing method for accurate traffic speed estimation with floating car
  data. \emph{Transportation Research Part C: Emerging Technologies}, Vol.~85,
  2017{\natexlab{c}}, pp. 644--663.

\bibitem[{Vlahogianni et~al.(2014)Vlahogianni, Karlaftis, and
  Golias}]{Vlahogianni.2014}
Vlahogianni, E.~I., M.~G. Karlaftis, and J.~C. Golias, {Short-term traffic
  forecasting: Where we are and where we're going}. \emph{{Transportation
  Research Part C: Emerging Technologies}}, Vol.~43, 2014, pp. 3--19.

\bibitem[{Bae et~al.(2018)Bae, Kim, Lim, Liu, Han, and
  Freeze}]{CokrigingImputation}
Bae, B., H.~Kim, H.~Lim, Y.~Liu, L.~D. Han, and P.~B. Freeze, Missing data
  imputation for traffic flow speed using spatio-temporal cokriging.
  \emph{Transportation Research Part C: Emerging Technologies}, Vol.~88, 2018,
  pp. 124 -- 139.

\bibitem[{Tan et~al.(2013)Tan, Feng, Feng, Wang, Zhang, and
  Li}]{TensorBasedCompletion}
Tan, H., G.~Feng, J.~Feng, W.~Wang, Y.-J. Zhang, and F.~Li, A tensor-based
  method for missing traffic data completion. \emph{Transportation Research
  Part C: Emerging Technologies}, Vol.~28, 2013, pp. 15 -- 27, euro
  Transportation: selected paper from the EWGT Meeting, Padova, September 2009.

\bibitem[{Ran et~al.(2016)Ran, Tan, Wu, and Jin}]{TensorBasedCompletion2}
Ran, B., H.~Tan, Y.~Wu, and P.~J. Jin, Tensor based missing traffic data
  completion with spatial–temporal correlation. \emph{Physica A: Statistical
  Mechanics and its Applications}, Vol. 446, 2016, pp. 54 -- 63.

\bibitem[{Chen et~al.(2019)Chen, He, and Sun}]{DataImputationTensor}
Chen, X., Z.~He, and L.~Sun, A Bayesian tensor decomposition approach for
  spatiotemporal traffic data imputation. \emph{Transportation Research Part C:
  Emerging Technologies}, Vol.~98, 2019, pp. 73 -- 84.

\bibitem[{{Li} et~al.(2019){Li}, {Zhang}, {Wang}, and
  {Ran}}]{ImputationMultiViewLearning}
{Li}, L., J.~{Zhang}, Y.~{Wang}, and B.~{Ran}, Missing Value Imputation for
  Traffic-Related Time Series Data Based on a Multi-View Learning Method.
  \emph{IEEE Transactions on Intelligent Transportation Systems}, Vol.~20,
  No.~8, 2019, pp. 2933--2943.

\bibitem[{Duan et~al.(2016)Duan, Lv, Liu, and Wang}]{DeepLearningImputation}
Duan, Y., Y.~Lv, Y.-L. Liu, and F.-Y. Wang, An efficient realization of deep
  learning for traffic data imputation. \emph{Transportation Research Part C:
  Emerging Technologies}, Vol.~72, 2016, pp. 168 -- 181.

\bibitem[{Li et~al.(2020)Li, Du, Wang, Qin, and Tan}]{DeepLImputationAppl}
Li, L., B.~Du, Y.~Wang, L.~Qin, and H.~Tan, Estimation of missing values in
  heterogeneous traffic data: Application of multimodal deep learning model.
  \emph{Knowledge-Based Systems}, Vol. 194, 2020, p. 105592.

\bibitem[{{Zhuang} et~al.(2019){Zhuang}, {Ke}, and {Wang}}]{DataImputationCNN}
{Zhuang}, Y., R.~{Ke}, and Y.~{Wang}, Innovative method for traffic data
  imputation based on convolutional neural network. \emph{IET Intelligent
  Transport Systems}, Vol.~13, No.~4, 2019, pp. 605--613.

\bibitem[{{Benkraouda} et~al.(2020){Benkraouda}, {Thodi}, {Yeo}, {Menéndez},
  and {Jabari}}]{deepCNNtrafficdata}
{Benkraouda}, O., B.~T. {Thodi}, H.~{Yeo}, M.~{Menéndez}, and S.~E. {Jabari},
  Traffic Data Imputation Using Deep Convolutional Neural Networks. \emph{IEEE
  Access}, Vol.~8, 2020, pp. 104740--104752.

\bibitem[{Helbing et~al.(2009{\natexlab{a}})Helbing, Treiber, Kesting, and
  Sch\"onhof}]{Helbing.2009}
Helbing, D., M.~Treiber, A.~Kesting, and M.~Sch\"onhof, Theoretical vs.
  Empirical Classification and Prediction of Congested Traffic States.
  \emph{The European Physical Journal B}, Vol.~69, 2009{\natexlab{a}}, pp.
  583--598.

\bibitem[{Thodi et~al.(2021)Thodi, Khan, Jabari, and
  Menendez}]{thodi2021incorporating}
Thodi, B.~T., Z.~S. Khan, S.~E. Jabari, and M.~Menendez, \emph{Incorporating
  Kinematic Wave Theory into a Deep Learning Method for High-Resolution Traffic
  Speed Estimation}, 2021.

\bibitem[{{Huang} and {Agarwal}(2020)}]{pinn_agarwal}
{Huang}, J. and S.~{Agarwal}, Physics Informed Deep Learning for Traffic State
  Estimation, 2020, pp. 1--6.

\bibitem[{Liu et~al.(2020)Liu, Barreau, Cicic, and
  Johansson}]{liu2020learningbased}
Liu, J., M.~Barreau, M.~Cicic, and K.~H. Johansson, \emph{Learning-based
  Traffic State Reconstruction using Probe Vehicles}, 2020.

\bibitem[{Shi et~al.(2021)Shi, Mo, Huang, Di, and Du}]{shi2021physicsinformed}
Shi, R., Z.~Mo, K.~Huang, X.~Di, and Q.~Du, Physics-Informed Deep Learning for
  Traffic State Estimation, 2021.

\bibitem[{Goodfellow et~al.(2016)Goodfellow, Bengio, Courville, and
  Bengio}]{goodfellow2016deep}
Goodfellow, I., Y.~Bengio, A.~Courville, and Y.~Bengio, \emph{Deep learning},
  Vol.~1. MIT press Cambridge, 2016.

\bibitem[{Long et~al.(2015)Long, Shelhamer, and Darrell}]{long2015fully}
Long, J., E.~Shelhamer, and T.~Darrell, \emph{Fully Convolutional Networks for
  Semantic Segmentation}, 2015.

\bibitem[{Chai et~al.(2020)Chai, Gu, Li, Duan, Hu, and Lin}]{Chai2020}
Chai, X., H.~Gu, F.~Li, H.~Duan, X.~Hu, and K.~Lin, Deep learning for
  irregularly and regularly missing data reconstruction. \emph{Scientific
  Reports}, Vol.~10, No.~1, 2020, p. 3302.

\bibitem[{Ghodrati et~al.(2019)Ghodrati, Shao, Bydder, Zhou, Yin, Nguyen, Yang,
  and Hu}]{QIMS29735}
Ghodrati, V., J.~Shao, M.~Bydder, Z.~Zhou, W.~Yin, K.-L. Nguyen, Y.~Yang, and
  P.~Hu, MR image reconstruction using deep learning: evaluation of network
  structure and loss functions. \emph{Quantitative Imaging in Medicine and
  Surgery}, Vol.~9, No.~9, 2019.

\bibitem[{Ioffe and Szegedy(2015)}]{batchnorm}
Ioffe, S. and C.~Szegedy, Batch Normalization: Accelerating Deep Network
  Training by Reducing Internal Covariate Shift. \emph{CoRR}, Vol.
  abs/1502.03167, 2015.

\bibitem[{Zhang(2019)}]{zhang_shiftinvariance}
Zhang, R., \emph{Making Convolutional Networks Shift-Invariant Again}, 2019.

\bibitem[{Kingma and Ba(2017)}]{kingma2017adam}
Kingma, D.~P. and J.~Ba, \emph{Adam: A Method for Stochastic Optimization},
  2017.

\bibitem[{Glorot and Bengio(2010)}]{glorot2010understanding}
Glorot, X. and Y.~Bengio, Understanding the difficulty of training deep
  feedforward neural networks. In \emph{Proceedings of the thirteenth
  international conference on artificial intelligence and statistics}, 2010,
  pp. 249--256.

\bibitem[{Srivastava and Geroliminis(2013)}]{geroliminis.capacitydrop}
Srivastava, A. and N.~Geroliminis, Empirical observations of capacity drop in
  freeway merges with ramp control and integration in a first-order model.
  \emph{Transportation Research Part C: Emerging Technologies}, Vol.~30, 2013,
  pp. 161 -- 177.

\bibitem[{Saberi and Mahmassani(2013)}]{Saberi.2013}
Saberi, M. and H.~S. Mahmassani, {Empirical Characterization and Interpretation
  of Hysteresis and Capacity Drop Phenomena in Freeway Networks}. \emph{{92th
  Annual Meeting of the Transportation Research Board}}, 2013.

\bibitem[{Chen et~al.(2014)Chen, Ahn, Laval, and
  Zheng}]{trafficOsciAndCapacityDrop}
Chen, D., S.~Ahn, J.~Laval, and Z.~Zheng, On the periodicity of traffic
  oscillations and capacity drop: The role of driver characteristics.
  \emph{Transportation Research Part B: Methodological}, Vol.~59, 2014, pp. 117
  -- 136.

\bibitem[{Treiber and Helbing(2002)}]{Treiber.2002}
Treiber, M. and D.~Helbing, Reconstructing the Spatio-Temporal Traffic Dynamics
  from Stationary Detector Data. \emph{Cooper@tive Tr@nsport@tion Dyn@mics},
  Vol.~1, 2002, pp. 3.1--3.24.

\bibitem[{Treiber et~al.(2011)Treiber, Kesting, and Wilson}]{Treiber.2011}
Treiber, M., A.~Kesting, and R.~E. Wilson, Reconstructing the Traffic State by
  Fusion of Heterogeneous Data. \emph{Computer-Aided Civil and Infrastructure
  Engineering}, Vol.~26, 2011, pp. 408--419.

\bibitem[{Schreiter et~al.(2010)Schreiter, {van Lint}, Treiber, and
  Hoogendoorn}]{Schreiter.2010}
Schreiter, T., H.~{van Lint}, M.~Treiber, and S.~Hoogendoorn, {Two Fast
  Implementations of the Adaptive Smoothing Method Used in Highway Traffic
  State Estimation}. \emph{{Intelligent Transportation Systems (ITSC)}}, Vol.
  13th International IEEE Conference on, 2010, pp. 1202--1208.

\bibitem[{Kessler et~al.(2018)Kessler, Huber, Kesting, and
  Bogenberger}]{Kessler.2018.TRB}
Kessler, L., G.~Huber, A.~Kesting, and K.~Bogenberger, Comparison of
  floating-car based speed data with stationary detector data. \emph{97th
  Annual Meeting of the Transportation Research Board}, 2018.

\bibitem[{Rempe et~al.(2016{\natexlab{b}})Rempe, Franeck, Fastenrath, and
  Bogenberger}]{Rempe.2016}
Rempe, F., P.~Franeck, U.~Fastenrath, and K.~Bogenberger, {Online Freeway
  Traffic Estimation with Real Floating Car Data}. \emph{{Intelligent
  Transportation Systems (ITSC)}}, 2016{\natexlab{b}}, pp. 1838--1843.

\bibitem[{{Chen} et~al.(2019){Chen}, {Zhang}, {Li}, and
  {Li}}]{asm_rolling_smoothing}
{Chen}, X., S.~{Zhang}, L.~{Li}, and L.~{Li}, Adaptive Rolling Smoothing With
  Heterogeneous Data for Traffic State Estimation and Prediction. \emph{IEEE
  Transactions on Intelligent Transportation Systems}, Vol.~20, No.~4, 2019,
  pp. 1247--1258.

\bibitem[{van Lint(2010)}]{Lint.2010}
van Lint, H., Empirical Evaluation of New Robust Travel Time Estimation
  Algorithms. \emph{Transportation Research Record}, Vol. 2160, No.~1, 2010,
  pp. 50--59.

\bibitem[{Bishop(2013)}]{Bishop.2013}
Bishop, C.~M., \emph{{Pattern recognition and machine learning}}. {Information
  science and statistics}, Springer, New York [u.a.], 11th ed., 2013.

\bibitem[{Helbing et~al.(2009{\natexlab{b}})Helbing, Treiber, Kesting, and
  Schoenhof}]{Helbing.}
Helbing, D., M.~Treiber, A.~Kesting, and M.~Schoenhof, {Theoretical vs.
  Empirical Classification and Prediction of Congested Traffic States}.
  \emph{{The European Physical Journal}}, Vol. 2009., No. B 69,
  2009{\natexlab{b}}, pp. 583--598.

\bibitem[{Yoon et~al.(2018)Yoon, Jordon, and van~der Schaar}]{gain_traffic}
Yoon, J., J.~Jordon, and M.~van~der Schaar, GAIN: Missing Data Imputation using
  Generative Adversarial Nets, 2018.

\end{thebibliography}
\end{document}